\documentclass{article}

\usepackage{arxiv}

\usepackage{graphicx}%
\usepackage{amsmath,amssymb,amsfonts}%
\usepackage{amsthm}%
\usepackage{mathrsfs}%
\usepackage[title]{appendix}%
\usepackage[table,x11names]{xcolor}
\usepackage{textcomp}%
\usepackage{manyfoot}%
\usepackage{booktabs}%
\usepackage{listings}%
\usepackage{bm}
\usepackage{bbm}
\usepackage{tabularx}
\usepackage{multirow}
\usepackage{multicol}
\usepackage{xspace}
\usepackage[acronym, shortcuts]{glossaries}
\usepackage{colortbl}
\usepackage{transparent}
\usepackage{hyperref}
\usepackage{tcolorbox}
\usepackage{cancel}
\usepackage{orcidlink}

\title{Multi-modal Co-learning for Earth Observation: Enhancing single-modality models via modality collaboration}

\date{} 			

\author{ 
    Francisco Mena$^{1,2}$\href{https://orcid.org/0000-0002-5004-6571}{\includegraphics[scale=0.06]{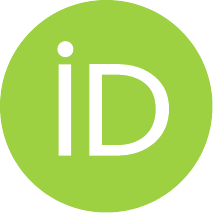}\hspace{1mm}},
	Dino Ienco$^{3,5}$\href{https://orcid.org/0000-0002-8736-3132}{\includegraphics[scale=0.06]{imgs/orcid.pdf}\hspace{1mm}}, 
	Cassio F. Dantas$^{3,5}$\href{https://orcid.org/0000-0002-1934-0625}{\includegraphics[scale=0.06]{imgs/orcid.pdf}\hspace{1mm}} 
    Roberto Interdonato$^{4,5}$\href{https://orcid.org/0000-0002-0536-6277}{\includegraphics[scale=0.06]{imgs/orcid.pdf}\hspace{1mm}} 
    Andreas Dengel$^{1,2}$\href{https://orcid.org/0000-0002-6100-8255}{\includegraphics[scale=0.06]{imgs/orcid.pdf}\hspace{1mm}}
    \\
	$^1$Department of Computer Science,	University of Kaiserslautern-Landau (RPTU), Kaiserslautern, Germany\\
	$^2$SDS, German Research Center for Artificial Intelligence (DFKI), Kaiserslautern, Germany\\
    $^3$INRAE, UMR TETIS, University of Montpellier, Montpellier, France\\
    $^4$CIRAD, UMR TETIS, University of Montpellier, Montpellier, France\\
    $^5$INRIA, EVERGREEN, University of Montpellier, Montpellier, France\\
	\texttt{f.menat@rptu.de} \\
}

\renewcommand{\undertitle}{Accepted at Machine Learning journal}
\renewcommand{\shorttitle}{Multi-modal Co-learning for Earth Observation}
\renewcommand{\headeright}{Mena et al.}

\newcommand*\citep[1]{\cite{#1}}

\newcommand{\mat}[1]{\mathbf{\uppercase{#1}}} 
\renewcommand{\vec}[1]{{\mathbf{\lowercase{#1}}} }

\newcommand{\letterfunc}[1]{\mathcal{#1}} 
\newcommand{\set}[1]{\mathbb{#1}}

\newcolumntype{L}{>{\raggedright\arraybackslash}X}
\newcolumntype{C}{>{\centering\arraybackslash}X}
\newcolumntype{R}{>{\raggedleft\arraybackslash}X}
\newcommand{\best}[1]{\pdfliteral direct {2 Tr 0.3 w}#1\pdfliteral direct {0 Tr 0 w}} 
\newcommand{\secondbest}[1]{{\underline{{#1}}}}

\newacronym{eo}{EO}{Earth Observation}
\newacronym{sits}{SITS}{Satellite Image Time Series}
\newacronym{vhsr}{VHSR}{Very High Spatial Resolution}
\newacronym{f1}{F1}{Weighted F1}
\newacronym{r2}{$R^2$}{Coefficient of Determination}
\newacronym{lfmc}{LFMC}{Live Fuel Moisture Content}
\newacronym{cropB}{CropH-b}{CropHarvest binary}
\newacronym{cropM}{CropH-m}{CropHarvest multi}
\newacronym{treesat}{{TSAITS}}{TreeSatAI-Time-Series}
\newacronym{mlp}{MLP}{Multi-Layer Perceptron}
\newacronym{tempcnn}{TempCNN}{Temporal CNN}
\newacronym{our}{MDiCo}{Multi-modal Disentanglement for Co-learning} 

\newcommand{\ntext}[1]{#1}

\begin{document}
\maketitle

\begin{abstract}
Multi-modal co-learning is emerging as an effective paradigm in machine learning, enabling models to collaboratively learn from different modalities \ntext{to enhance} single-modality predictions. 
Earth Observation (EO) represents a quintessential domain for multi-modal data analysis, wherein diverse remote sensors collect data to sense our planet. This unprecedented volume of data introduces novel challenges.
Specifically, the access to the same sensor modalities at both training and inference stages becomes increasingly complex based on real-world constraints affecting remote sensing platforms. 
In this context, multi-modal co-learning presents a promising strategy to leverage the vast amount of sensor-derived data available at the training stage to improve single-modality models for inference-time deployment. Most current research efforts focus on designing customized solutions for either particular downstream tasks or specific modalities available at the inference stage.
To address this, we propose a novel multi-modal co-learning framework capable of generalizing across various tasks without targeting a specific modality for inference. 
Our approach combines contrastive and \ntext{modality discriminative} learning together to guide single-modality models to \ntext{structure the internal model manifold into modality-shared and modality-specific} information.
We evaluate our framework on four EO benchmarks spanning classification and regression tasks across different sensor modalities, where only one of the modalities available during training is accessible at inference time. 
Our results demonstrate consistent predictive improvements over state-of-the-art approaches from the recent machine learning and computer vision literature, as well as EO-specific methods.
The obtained findings validate our framework in the single-modality inference scenarios across a diverse range of EO applications.
\end{abstract}


\section{Introduction} \label{sec:intro}

The collaborative learning paradigm, named co-learning, has been largely studied in the machine learning field \cite{blum1998combining,han2018co}. The objective is to have multiple models (or layers) that share knowledge and cooperate \ntext{with} each other to improve their performance. This learning paradigm has been adopted for domain adaptation \cite{ganin2015unsupervised,obrenovic2023learning}, federated learning \cite{carrascosa2022co},  learning with noisy labels \cite{han2018co}, and knowledge distillation \cite{ienco2024discom}. Recently, with the increasing availability of multi-modal data acquired through a plethora of different platforms and sensors, co-learning can play an even more crucial role~\cite{rahate2022multimodal}.

\begin{figure}[!t]
	\centering
	\includegraphics[width=\textwidth]{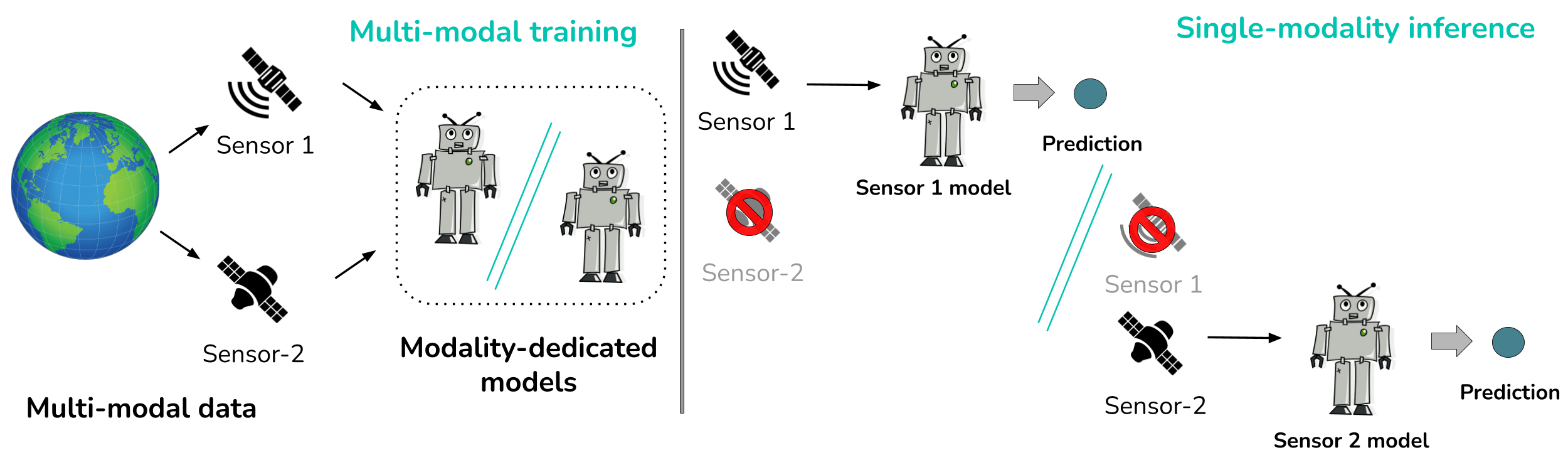}
	\caption{Illustration of multi-modal co-learning. The multi-modal data is available for training, but single-modality data is accessible for inference as an all-but-one missing modality scenario.}  \label{fig:colearn}
\end{figure}

Multi-modal data describing the same phenomena of interest have been used for different purposes in the literature. Multi-modal learning targets the effective exploitation of multiple data modalities for improving model performance and supporting informed decisions \cite{baltruvsaitis2018multimodal}.
\ntext{Recent strategies} rely on data fusion mechanisms to improve accuracy or to enhance synergy between modalities via self-supervision. 
However, we consider multi-modal data under a co-learning paradigm in this work, focusing on the scenario where only one of the modalities available at training is accessible at inference time \cite{wu2024deep}, referred to as \textit{all-but-one missing modality} \cite{rahate2022multimodal}. More precisely, we do not make any assumption on which modality is accessible at inference time, as shown in Fig.~\ref{fig:colearn}.
\ntext{To address this point}, Andrew et al. \cite{andrew2013deep} learn a shared feature space between modalities by maximizing the cross-covariance, while Zhang et al. \cite{zhang2024multimodal} share the last linear layer among single-modality models to force sharing information between modalities.
The benefit of learning from multi-modal data has been demonstrated empirically for single-modality inference, as well as studied theoretically by Zadah et al. \cite{zadeh2020foundations}. 
Thus, multi-modal co-learning has been applied to different domains where only a subset of the modalities is accessible at the inference stage, such as in \gls{eo}.

Nowadays, the \gls{eo} domain is characterized by a vast amount of heterogeneous multi-modal data.
Thanks to the advances in instruments and technology, numerous satellites are constantly sensing the Earth's surface \cite{camps2021deep}.
Such multi-modal data requires advanced methods to take advantage of the carried complementary information~\cite{mena2024common}.
This is because the collected sensor data is diverse and heterogeneous due to: different acquisition modes (e.g. optical and radar), spectral characteristics (e.g. RGB and multi-spectral), and resolutions (e.g. spatial and temporal).
Thus, \gls{eo} sensor data go beyond standard benchmarks that mainly cover natural images in the standard computer vision domain \cite{rolf2024mission}, limiting the applicability of mainstream machine learning and vision models to the \gls{eo} domain.
Moreover, accessing the same set of sensor data during both the training and inference stages could be infeasible in scenarios characterized by operational constraints, resulting in missing modality scenarios \cite{wu2024deep}.

The lack of systematically available sensor data covering the same region over the same period is an inherent problem in the \gls{eo} domain \cite{shen2015missing}.
This is because sensor data collection occurs under operational constraints in real-world environments, where geographical extent, weather acquisition conditions, deployment costs, and sensor failures may affect its consistent and systematic availability. Thus, sensor modalities in multi-modal scenarios can be partially or entirely missing during inference. For instance, the Landsat 7 ETM+ SLC-off problem after 2003 \cite{markham2004landsat}, the Sentinel-1b satellite that stopped operating at the end of 2021 \cite{potin2022status}, and the NAIP satellite that operates only in the United States.
Thus, enhanced collaboration between modalities available in the training stage is essential if modalities are missing at inference.

Recently, multi-modal co-learning has been used in the \gls{eo} domain to face the problem of missing modality during inference \cite{kieu2024multimodal}, as shown in Fig.~\ref{fig:colearn}.
\ntext{However, most of the co-learning exploration} has been limited to frameworks tailored for a predefined missing modality during inference, named \textit{dedicated training}~\cite{wu2024deep}. 
In this way, training aims to use additional modalities only as a support for the main target modality.
For instance, Kampffmeyer et al. \cite{kampffmeyer2018urban} use a hallucination branch from an optical modality to simulate the depth modality that is expected to be missing, \ntext{while MMEarth model} \cite{nedungadi2024mmearth} \ntext{reconstructs various other sensor modalities from a specific one}.
Since \ntext{these approaches define in advance the modalities that will be missing during inference, they are not general enough to cases when other modalities} might be absent.
\ntext{Conversely, \textit{non-dedicated training} avoids making in advance the firm choice on the modality available during inference. For instance,} Zheng et al. \cite{zheng2021deep} use a common model generator to instantiate single-modality models, while Mena et al. \cite{mena2024maclean} share weights of the last layers between modality-dedicated models.  \ntext{Similarly to this latter family of approaches, we address the missing modality problem with a non-dedicated training strategy.}

In this manuscript, we propose a multi-modal co-learning framework to handle the problem of all-but-one missing modality occurring at inference time with \gls{eo} data. Our framework focuses on boosting the cross-modal knowledge transfer between heterogeneous sensor modalities \ntext{at the feature-level}. 
\ntext{To this end,} we use modality-dedicated encoders to extract common and unique information. 
\ntext{More precisely, we enforce a disentanglement of three feature spaces (shared, specific, and unused) per modality that are guided by several loss functions, named} {\gls{our}}.
Unlike \ntext{prior co-learning approaches} in the \gls{eo} domain, \ntext{which often focus} on a single task, e.g. land-cover classification, or design frameworks tightly coupled to specific data modalities, we propose a framework that is task-agnostic and adaptable to the modality available at inference time.
\ntext{To achieve this flexibility, we incorporate dedicated prediction heads trained explicitly to predict the downstream task from the shared and specific feature representations per-modality, enabling inference from any of the modality available at the training stage.}

To comprehensively assess the behavior of our framework, we compare several recent state-of-the-art approaches from the general domain of machine learning and computer vision, as well as four recent methods especially tailored for the \gls{eo} domain.
Our experimental assessment covers four multi-modal \gls{eo} benchmarks, featured by different combinations of sensor modalities, spanning several downstream tasks such as binary, multi-class, and multi-label classification, as well as regression.
The results prove the quality of \gls{our} over the competing approaches across the different downstream tasks and single-modality scenarios at the inference stage.
The systematic improvement exhibited by \gls{our}, in all validation scenarios, positions our framework as an effective solution for multi-modal co-learning where all-but-one missing modality scenarios arise at the inference stage. Our code and related datasets are available at \url{https://github.com/fmenat/MDiCo}.

This manuscript is organized as follows: The related literature on co-learning, multi-modal data, and \gls{eo} is described in Sec.~\ref{sec:related}. Sec.~\ref{sec:method} introduces our multi-modal co-learning framework. The experimental evaluation and the related findings are reported and discussed in Sec.~\ref{sec:exp}, while Sec.~\ref{sec:conclu} draws the conclusions of our work.

\section{Related work} \label{sec:related}

\subsection{Multi-modal learning and missing modalities}

\ntext{
Multi-modal learning has proven to be effective in enhancing model performance and generalization. This paradigm involves training deep learning models on multiple data modalities simultaneously {\cite{baltruvsaitis2018multimodal}}, such as images, texts, and sensor signals, enabling models to leverage complementary information across heterogeneous data. 
Beyond performance improvements, research works have explored how multi-modal approaches influence the learning of distinct feature representations per modality.  To extract common (or shared) feature representations across modalities, various loss functions have been employed, including correlation maximization {\cite{andrew2013deep}}, contrastive learning {\cite{yuan2021multimodal}}, and weight-sharing mechanisms across modality-specific models {\cite{ngiam2011multimodal}}. 
For instance, Poklukar et al. {\cite{poklukar2022geometric}} demonstrate that shared features derived via contrastive loss remain robust to missing all-but-one modality.
Other approaches aim to disentangle shared and specific features jointly by combining multiple loss functions.
The MISA model {\cite{hazarika2020misa}} learns shared features through both weight-sharing and minimizing inter-modality distances, while specific features are extracted using separate encoders and enforced by minimizing similarity across modalities.
In contrast, the ShaSpec model {\cite{wang2023multi}} employs a domain discriminator to learn modality-specific features. 
However, multi-modal learning can still suffer from the missing modality problem.

The challenge of missing modalities at inference time has been addressed through various strategies in the literature {\cite{wu2024deep}}. One common approach involves simulating missing modalities during training, as demonstrated in works such as {\cite{wang2023multi}} and {\cite{mena2025maug}}. 
A different strategy, explored by Choi et al. {\cite{choi2019embracenet}}, involves randomly selecting a single modality during training, at each feature dimension, to encourage robustness. 
Furthermore, knowledge distillation and self-distillation frameworks have been employed, where a full-modality teacher guides student models trained with incomplete modalities, as introduced in the approaches by McKinzie et al. {\cite{mckinzie2023robustness}} and Lin \& Hu {\cite{lin2023missmodal}}.
}

\subsection{Multi-modal co-learning}

Co-learning has been used in diverse fields and applications by having multiple models that cooperate among them. 
In domain adaptation, Ganin et al. \cite{ganin2015unsupervised} introduce per-domain models with shared encoders and a domain classifier for image classification, while Obrenovic et al. \cite{obrenovic2023learning} introduce a similar approach by sharing only the middle layers among per-domain models in a heterogeneous domain adaptation \ntext{setting}.
In learning with noisy labels, co-teaching \cite{han2018co} uses two models trained simultaneously that \ntext{supervise each other}, selecting samples with potential clean labels. Similarly, MentorNet \cite{jiang2018mentornet} uses a teacher network to guide the training via selecting reliable instances to automate curriculum learning.
Under the lens of mutual distillation, Zhang et al. \cite{zhang2018deep} introduce a framework with two identical models that are cross-guided based on predicted probabilities matching for image classification tasks.

Co-learning has emerged in the multi-modal setting to handle noisy modalities and unreliable label scenarios \cite{rahate2022multimodal}.
The co-learning \ntext{process} between single-modality models can be done at the feature-, decision-, or model-level. For instance, learning a shared feature space between modalities has been implemented via deep learning models as a feature-based co-learning strategy \cite{andrew2013deep} and for cross-modal distillation \cite{ienco2024discom}.
Recently, it has been used in contrastive learning for self-supervision \cite{yuan2021multimodal}.
Moreover, decision-based co-learning, i.e. knowledge transfer between single-modality model predictions, has been used in a (multi-modal) semi-supervised setting \cite{qiao2018deep} and for mutual distillation \cite{black2024multi}.
Furthermore, the model-based co-learning involves sharing layers among the single-modality models. For instance, Zhang et al. \cite{zhang2024multimodal} use a shared prediction head among single-modality models with orthogonal directions in the gradient \ntext{trajectory}. On the other hand, Zadah et al. \cite{zadeh2020foundations} theoretically show that learning from multi-modal information is better even if testing scenarios have a single-modality setting, i.e. the model can benefit from additional information available only during training.

\subsection{Multi-modal co-learning in EO}
Multi-modal data has been crucial in the \gls{eo} domain to analyze complex phenomena on Earth \cite{camps2021deep}.
This is because sensors capture different information about the Earth's surface, complementing individual observations, such as optical and radar data.
Thus, most of the research leveraging multi-modal data focuses on data fusion to enhance model accuracy in different applications \cite{tuia2021toward}.
One example is the data fusion contest hosted each year by the Geoscience and Remote Sensing Society \cite{persello2023,persello2024}. Moreover, Mena et al. \cite{mena2024common} discuss different fusion strategies used in the \gls{eo} domain and underline the tendency to design models (and their combinations) with increasing levels of complexity. However, efforts have focused on designing simple models employing cross-distillation strategies involving a multi-modal teacher and single-modal students \cite{bakalos2024segmentation}.
For instance, Pande et al. \cite{pande2019adversarial} use an adversarial approach with a hallucination network on multi-spectral and panchromatic images. 

Multi-modal co-learning has been effectively used to handle missing sensor modalities in the \gls{eo} domain \cite{kieu2024multimodal}. The most common scenario is feature-based co-learning, where features are forced to be similar, such as with hallucination networks \cite{kampffmeyer2018urban}, cross-modal retrieval \cite{chen2020deep}, and contrastive learning \cite{dantas2024reuse}.
Recently, contrastive learning has been leveraged for multi-modal model pretraining, such as in Heidler et al. \cite{heidler2023self} and OmniSat \cite{astruc2025omnisat}.
On the model-based co-learning aspect, Hong et al. \cite{hong2019cospace} introduce the sharing of prediction heads between hyper- and multi-spectral optical image models. Then, Zheng et al. \cite{zheng2021deep} propose a meta-model that generates the parameters of the single-modality models in an optical-radar sensor setting.
Moreover, Xie et al. \cite{xie2023co} use a co-training strategy to obtain pseudo-labels from one modality to another in a semi-supervised setting with point cloud data and optical images. \ntext{Another family of} strategies available in the literature relies on cross-reconstruction. For instance, MMEarth \cite{nedungadi2024mmearth} reconstructs various sensor modalities from a specific one. On the same track, Xiong et al. \cite{xiong2024} share layers between modalities to set up a reconstruction task.

Previous co-learning works in the \gls{eo} domain \ntext{commonly} focus on single tasks and design specific methods for the used modalities \cite{zheng2021deep,xie2023co} or assume a modality-dedicated training {\cite{kampffmeyer2018urban,pande2019adversarial,bakalos2024segmentation,nedungadi2024mmearth}}.
In contrast to these works, we introduce a general multi-modal co-learning approach that is flexible enough to be used for any available modality across a diverse set of downstream tasks.

\section{Method} \label{sec:method}

\subsection{Notation}
Let us consider the multi-modal co-learning under a supervised setting as follows.
There are $N$ training samples available with corresponding multi-modal data and ground truth information.
Without loss of generality, we consider a two-modality setup as $\set{D} = \{ \mat{X}_1^{(i)}, \mat{X}_2^{(i)}, y^{(i)} \}_{i=1}^N$.
During inference, only one of the modalities is available, either $\mat{X}_1$ or $\mat{X}_2$, as illustrated in Fig.~\ref{fig:colearn}.
Each modality $\mat{X}_m$, with $m\in \{1,2\}$ provides per-modality specific information. Here, the objective is to derive single-modality models $\letterfunc{G}_m(\cdot)$, where each model is composed of an encoder and prediction head, that supply the prediction for test data, i.e. $\hat{y}_m^{(i)} = \letterfunc{G}_m(\mat{X}_m^{(i)})$.
For simplicity, we avoid the superscript of the sample index $i$ in the following.

\subsection{Framework description}

\begin{figure}[!t]
	\centering
	\includegraphics[width=\textwidth]{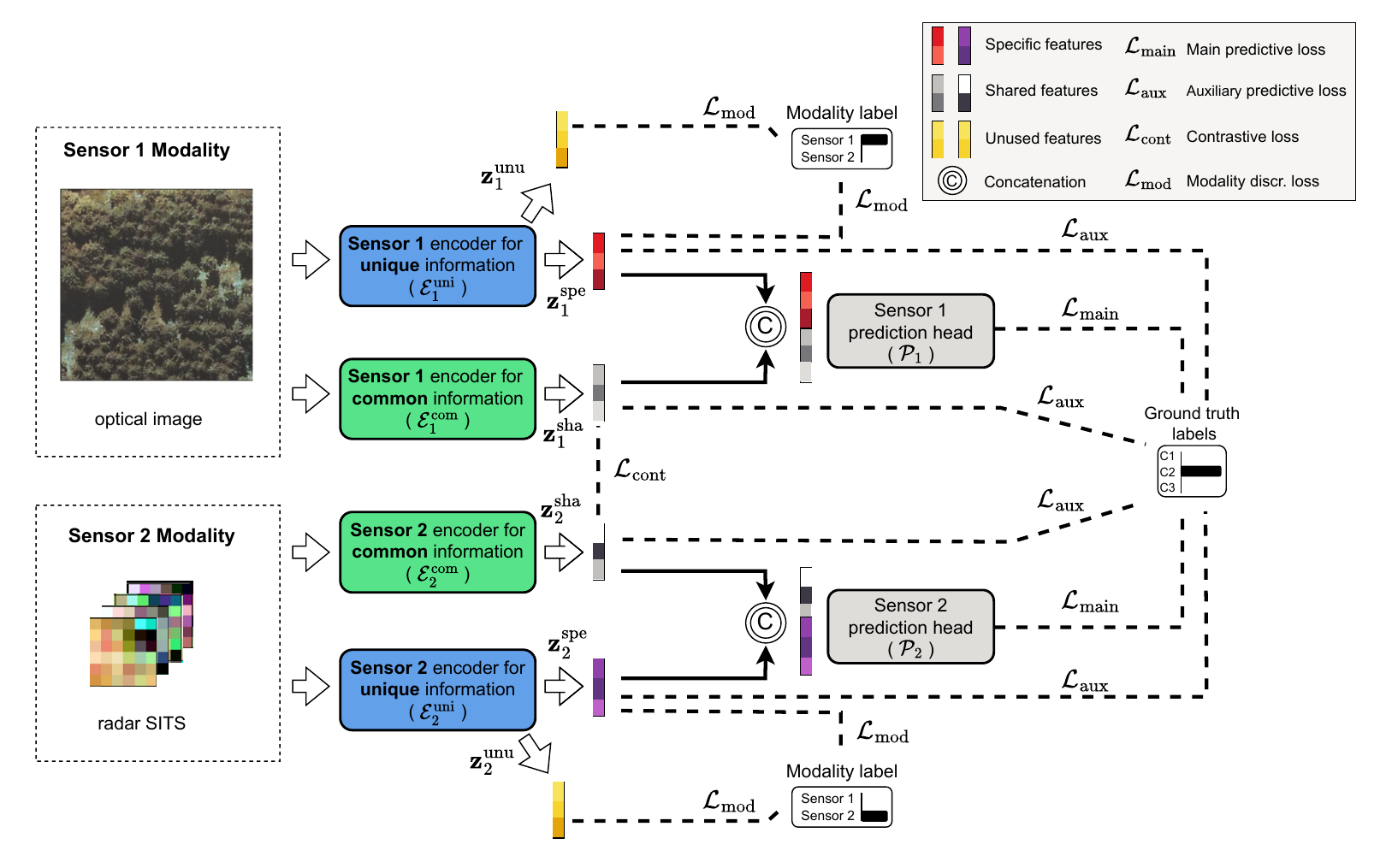}
	\caption{Illustration of our \gls{our} framework with shared, specific, and unused features. The main prediction per modality is illustrated in regular arrows, while the losses are shown with dashed lines.
	Two sensor modalities are shown: an optical image and a radar \acrfull{sits}.
	\label{fig:model}}  
\end{figure}

To address the problem of all-but-one missing modality at the inference stage (see Fig.~\ref{fig:colearn}), we introduce a framework that facilitates the collaboration between heterogeneous sensor modalities and improves the cross-modal learning \ntext{at the feature-level}.

\subsubsection{Overview} \label{sec:method:over}
Our framework, shown in Fig.~\ref{fig:model}, considers modality-dedicated encoders that extract three representations per modality. Two of these representations are used for the downstream task, considering the \ntext{modality-}shared and \ntext{modality-}specific feature space, while the third one, referred to as unused, is \ntext{successively discarded}.
\ntext{To disentangle these feature spaces and guide the co-learning process, we use multiple loss functions. 
Two of these losses are at feature-level while the remaining two are at prediction-level.}
Thus, our \textbf{{\acrfull{our}}} framework is task-agnostic, adopting a predictive loss function driven by each downstream task.

During inference, the target is predicted from each single-modality model based on the shared and specific features coming from the per-modality encoders.
Thus, our multi-modal formulation operates with arbitrary single-modality available at the inference stage.
This corresponds to a more general case than dedicated training in the literature \cite{kampffmeyer2018urban,bakalos2024segmentation}, which considers a predefined modality accessible at inference time.

\subsubsection{Shared, Specific and Unused Feature Spaces} \label{sec:method:spaces}

Consider per-modality encoders $\letterfunc{E}_m^{\text{com}}(\cdot)$, with $m\in \{1,2\}$, that extract common information between modalities. This common information 
is obtained as follows:
\begin{equation}
	\vec{z}_m^{\text{sha}} =  \letterfunc{E}_m^{\text{com}}\left( \mat{X}_m \right) \ \  \forall m \in \{1,2\}  ,
\end{equation}
with  $\vec{z}_m^{\text{sha}} \in \mathbb{R}^d$ the \textbf{shared features}.
To achieve the learning of shared features between modalities, we use a contrastive learning strategy (cf. Sec.~\ref{sec:method:loss:contr}).
Furthermore, consider modality-dedicated encoders $\letterfunc{E}_m^{\text{uni}}(\cdot)$ that extracts information that is unique to the modality $m$. This unique information can be separated into task-discriminative and non-discriminative, similar to \cite{ienco2024discom}, modeled as follows:
\begin{equation}
	\vec{z}_m^{\text{spe}}, \vec{z}_m^{\text{unu}} =  \letterfunc{E}_m^{\text{uni}} \left(\mat{X}_m\right) \ \  \forall m \in \{1,2\}  ,
\end{equation}
with  $\vec{z}_m^{\text{spe}}  \in \mathbb{R}^d$ the \textbf{specific features}, and $\vec{z}_m^{\text{unu}}  \in \mathbb{R}^d$ the \textbf{unused features}.
The guidance for learning  \ntext{this unique per-modality information} is achieved via a modality discriminator strategy discussed in Sec.~\ref{sec:method:loss:modality}. 
\ntext{The difference between these two feature spaces is that we adopt an explicit predictive loss over the specific features.}
We opt to discriminate the unused features in this unique information, based on the hypothesis that each modality could have noise that is unrelated to the downstream task.
Moreover, we did not assume any fixed shape for the per-modality encoders, $\letterfunc{E}_m^{\text{com}}$ and $\letterfunc{E}_m^{\text{uni}}$, as they tightly depend on the input data (e.g. image or time series).

\subsubsection{Per-Modality Prediction} \label{sec:method:pred}
We use a per-modality prediction head $\letterfunc{P}_m(\cdot)$ that takes the concatenation of the task-discriminative feature spaces, i.e. $\vec{z}_m^{\text{sha}}$ and $\vec{z}_m^{\text{spe}}$, and estimates the target value, expressed by
\begin{equation} \label{eq:per_modality:pred}
	\hat{y}_m = \letterfunc{P}_m \left( [ \vec{z}_m^{\text{sha}} \ || \ \vec{z}_m^{\text{spe}} ] \right) \ \  \forall m \in \{1,2\}  .
\end{equation}
In this way, the framework splits into single-modality models used for the single-modality prediction at inference time, i.e. $\letterfunc{G}_m = \{ \letterfunc{P}_m,\letterfunc{E}_m^{\text{com}}, \letterfunc{E}_m^{\text{uni}} \}$, given by
\begin{align}
	\hat{y}_m & = \letterfunc{G}_m \left( \mat{X}_m \right)  \\
	& = \letterfunc{P}_m \left( [ \letterfunc{E}_m^{\text{com}}\left( \mat{X}_m \right) \ || \ \letterfunc{E}_m^{\text{uni}} \left(\mat{X}_m\right)_{\text{spe}} ] \right) \ ,
\end{align}
where $\letterfunc{E}_m^{\text{uni}} \left(\mat{X}_m\right)_{\text{spe}}$ is the specific features in the unique information, i.e. $\vec{z}_m^{\text{spe}}$.

\subsection{Training losses} \label{sec:method:loss}
To disentangle the learning of the three per-modality feature spaces, we employ several loss functions. The resulting final loss function optimized during training is designed as follows:
\begin{equation} \label{eq:loss}
 	\letterfunc{L}_{\text{total}} =  \letterfunc{L}_{\text{main}} +  \letterfunc{L}_{\text{aux}} + \letterfunc{L}_{\text{cont}}  +  \letterfunc{L}_{\text{mod}}  \ . 
\end{equation}
\ntext{We use an unweighted sum of the loss functions in our {\gls{our}} framework (Eq.~{\eqref{eq:loss}})}. 
In the following, each of the loss functions is described.

\subsubsection{Main Predictive loss}\label{sec:method:loss:main}
As the main loss in our framework, we use a task-discriminative loss to guide the learning process of the per-modality shared and specific features. Considering the per-modality prediction $\hat{y}_m$ in Eq.~\eqref{eq:per_modality:pred},  with $m \in \{1,2\}$, and the associated ground truth $y$, the task-discriminative loss is computed as follows:

\begin{equation} \label{eq:loss:main}
	\letterfunc{L}_{\text{main}} = \frac{1}{2} \sum_{m \in \{1,2\}} \letterfunc{L}_{\text{pred}} \left( y, \hat{y}_m \right) ,
\end{equation}
with $\letterfunc{L}_{\text{pred}}(\cdot, \cdot)$ the cross-entropy loss function commonly used in multi-class classification tasks, i.e. $\letterfunc{L}_{\text{CE}}(p,q) = - \sum_k \mathbbm{1}(p=k) \log{q_k}$ with $\mathbbm{1}(\cdot)$ the indicator function, or the squared error in regression tasks, i.e. $\letterfunc{L}_{\text{SE}}(p,q) = (p-q)^2$.

\subsubsection{Auxiliary Predictive loss} \label{sec:method:loss:aux}
Additionally, we consider an auxiliary predictive loss to enhance the learning of task-discriminative information for both shared and specific per-modality features.
Here, we consider a single auxiliary prediction head $\letterfunc{P}^{\text{aux}}(\cdot)$  that predicts the target $\hat{y}^{\text{aux}}$ across modalities and that is used on both feature spaces, $ \vec{z}_m^{\text{sha}}$ and $\vec{z}_m^{\text{spe}}$  with $m \in \{1,2\}$.
Thus, the auxiliary predictive loss is computed as:
\begin{equation}
	\letterfunc{L}_{\text{aux}} = \frac{1}{2} \sum_{m \in \{1,2\}} \ \sum_{s \in \{\text{sha}, \text{spe}  \}} \letterfunc{L}_{\text{pred}} \left( y,\  \letterfunc{P}^{\text{aux}} \left( \vec{z}_m^{s} \right) \right) ,
\end{equation}
with $\letterfunc{L}_{\text{pred}}(\cdot, \cdot)$ the same loss function used in Eq.~\eqref{eq:loss:main}.

\subsubsection{Contrastive loss} \label{sec:method:loss:contr}
For learning the features shared among modalities, i.e. $\vec{z}_m^{\text{sha}}$ with $m \in \{1,2\}$, we use the contrastive learning \cite{chen2020simple} by forcing the paired features between modalities to be similar in the learned latent space. 
In this way, the shared features are guided to group themselves and be modality invariant.
We use the InfoNCE \cite{chen2020simple} criteria given by:
\begin{equation}\label{eq:contrasitve_sim}
    \letterfunc{L}_{\text{info}}( \vec{z}_1^{\text{sha}}, \vec{z}_2^{\text{sha}}; \tau ) = - \log{ \frac{ \exp{ \left( \letterfunc{S}( \vec{z}_1^{\text{sha}}, \vec{z}_2^{\text{sha}} ) / \tau \right)} }{ \sum\limits_{ \tilde{\vec{z}}^{\text{sha}}_2 \in \set{Z}_2  } \exp{ \left( \letterfunc{S}( \vec{z}_1^{\text{sha}} , \tilde{\vec{z}}_2^{\text{sha}} ) / \tau \right) }  } } \ ,
\end{equation}
with the cosine similarity as $\letterfunc{S}(a,b) = \langle a,b \rangle / (\Vert a \Vert_2 \!\cdot\! \Vert b \Vert_2)$ and $\langle \cdot, \cdot \rangle $ the inner product between two vectors,  $\tau >0 $ a temperature hyper-parameter that we set to $0.07$ following~\cite{dantas2024reuse}, and $\set{Z}_*$ the set of all other cross-modality pairs in the batch. 
Here, positive pairs (numerator in Eq.~\eqref{eq:contrasitve_sim}) are the two modalities from the same sample, while the negative ones (denominator in Eq.~\eqref{eq:contrasitve_sim}) are the complementary modality features that come from all other samples in the batch. 
Thus, we consider each modality $m \in \{1, 2\}$ to be the anchor in the contrastive loss, which is given by:
\begin{equation} \label{eq:contrastive}
	\letterfunc{L}_{\text{cont}} = \letterfunc{L}_{\text{info}}( \vec{z}_1^{\text{sha}}, \vec{z}_2^{\text{sha}} ; \tau)  + \letterfunc{L}_{\text{info}}( \vec{z}_2^{\text{sha}}, \vec{z}_1^{\text{sha}} ; \tau )  \ .
\end{equation}

\subsubsection{Modality Discriminant loss} \label{sec:method:loss:modality}
For learning the features proper to each modality, i.e. $\vec{z}_m^{\text{spe}}$ and $\vec{z}_m^{\text{unu}}$  with $m \in \{1,2\}$, we use an auxiliary classifier that discriminates from which modality the features are coming from.
To this end, we use two classifiers, one for the specific $\letterfunc{P}^{\text{spe}}(\cdot)$ and another for the unused $\letterfunc{P}^{\text{unu}}(\cdot)$ feature spaces, where the loss function is designed as:
\begin{equation}
	\letterfunc{L}_{\text{mod}} = \frac{1}{2} \sum_{m \in \{1,2\}} \  \sum_{s \in \{\text{spe}, \text{unu}  \}}  \letterfunc{L}_{\text{CE}} \left( m, \  \letterfunc{P}^{\text{s}} \left( \vec{z}_m^{\text{s}} \right)  \right) ,
\end{equation}
with $\letterfunc{L}_{\text{CE}}(\cdot, \cdot)$ the standard cross-entropy loss function.

\section{Experiments} \label{sec:exp}

We evaluate our framework against several approaches from recent machine learning, computer vision, and \gls{eo} literature.
The analysis encompasses four multi-modal {\gls{eo}} benchmarks covering classification (binary, multi-class, and multi-label) and regression tasks. 
\ntext{
Concretely, we first introduce the benchmarks (Sec.~{\ref{sec:data}}) and competing approaches (Sec.~{\ref{sec:competing}}). Then, we describe the evaluation protocol and setup (Sec.~{\ref{sec:setting}}).
Subsequently, we present and discuss the obtained results (Sec.~{\ref{sec:results}}).
To provide deeper insights, we examine individual components in our framework through ablation studies (Sec.~{\ref{sec:abla}} and~{\ref{sec:shaspe}}). 
Furthermore, we assess our framework using various encoder backbones (Sec.~{\ref{sec:enco}}). Finally, we analyze the internal representations learned by {\gls{our}} (Sec.~{\ref{sec:quali}}), and the evolution of the individual losses during training (Sec.~{\ref{sec:losses}}).
}

\subsection{Datasets} \label{sec:data}
For the binary and multi-class classification task, we consider a crop recognition problem by using the CropHarvest benchmark \cite{tseng2021crop}.
This benchmark contains samples around the globe between 2016 and 2021.
The benchmark contains two multi-temporal sensor modalities, corresponding to \gls{sits}, at a spatial resolution of $10 [m]$: optical
Sentinel-2 \gls{sits} (considering spectral bands and vegetation indices), and radar Sentinel-1 \gls{sits} (with polarization bands). Thus, we consider the crop/non-crop estimation at a specific region as a cropland (binary) classification task. This dataset, named \textbf{\gls{cropB}} has 69\,800 samples.
For multi-class classification, we use a subset of the CropHarvest benchmark including 29\,642 samples with crop-type labels associated. More precisely, it
covers ten different classes for a downstream crop-type classification task. This variant is named \textbf{\gls{cropM}}.

For the multi-label classification task, we consider a tree-species identification problem introduced in \cite{astruc2025omnisat}, referred to as \textbf{\gls{treesat}}.
It involves a multi-label classification task covering 15 tree species in Germany. This benchmark includes a multi-temporal and a mono-temporal modality: Sentinel-2 \gls{sits} at a $10 [m]$ spatial resolution (with multi-spectral bands), and an aerial (mono-temporal) image at a high spatial resolution of $0.2 [m]$ (with RGB and infrared bands).
This benchmark includes 38\,520 samples for training, 6\,810 for validation, and 5\,044 for testing, collected between 2017 and 2020.

For the regression task, we consider a moisture estimation task introduced in \cite{rao2020sar}, referred to as \textbf{\gls{lfmc}}.
It involves the prediction of vegetation water (moisture) content per dry biomass in the western US. For this task, we have access to two multi-temporal modalities at a spatial resolution of $250 [m]$: Landsat 8 \gls{sits} (with spectral bands and vegetation indices), and Sentinel-1 \gls{sits} (with polarization bands and indices). This dataset contains 2\,578 samples collected between 2015 and 2019.

A summary of the benchmark information with the corresponding modality characteristics and the data \ntext{format} is reported in Table~\ref{tab:data_descr}.

\begin{table}[!t]
	\centering
	\caption{Description of data modalities used in each dataset. The input \ntext{format} corresponds to (time-steps, features, height, width).}	\label{tab:data_descr}
	\begin{tabularx}{\textwidth}{c|ccCCc} \toprule
     	Dataset & Modality & Sensor type & Spatial resolution & Temporal resolution & Input shape \\
     	\midrule
     	\acrshort{cropB} \&  & Sentinel-1 & radar \gls{sits} & 10 $[m]$ & 1 per month & (12, 11, 1, 1) \\
     	\acrshort{cropM} & Sentinel-2 & optical \gls{sits} & 10 $[m]$ & 1 per month &  (12, 2, 1, 1) \\
     	\midrule
     	\acrshort{treesat} & Aerial & optical image & 0.2 $[m]$ &  None &  (1, 4, 320, 320)  \\
      	& Sentinel-2 & optical \gls{sits} & 10 $[m]$ & $\sim$10 per month &  (150, 12, 6, 6) \\
      	\midrule
      	\acrshort{lfmc}  & Sentinel-1 & radar \gls{sits} & 250 $[m]$ & 1 per month & (4, 3, 1, 1) \\
     	& Landsat 8 & optical \gls{sits} & 250 $[m]$ & 1 per month & (4, 8, 1, 1)\\
     	\bottomrule
	\end{tabularx}
\end{table}

\subsection{Competing methods} \label{sec:competing}
For the assessment of our framework, we consider several families of competitors. 
Using a dedicated training, we include models individually trained on each modality as single-modality baselines, named \textit{Individual}. In addition, we consider models fusing the multi-modal sensor data.

\ntext{Incorporating a non-dedicated training}, we select methods from the recent literature that are especially tailored for handling the different downstream tasks we validate on. From the field of multi-modal fusion with missing modalities, we include two methods: EmbraceNet \cite{choi2019embracenet}, a feature-level fusion model employing feature sampling as a merge function, and ShaSpec \cite{wang2023multi}, a feature-level fusion model using shared and specific features to handle missing modalities.
We include two methods from the multi-modal co-learning field:
DeCuR \cite{wang2023decur}, a self-supervised approach that learns common and unique features in single-modality models, DML \cite{zhang2018deep}, an ensemble using mutual distillation to match the predictions of single-modality models.
From the cross-modal distillation field, we include DisCoM-KD \cite{ienco2024discom}, a recent method using invariant, informative, and irrelevant features for the cross-modal learning of paired modalities.
Finally, we include four recent methods from the \gls{eo} literature, AnySat \cite{astruc2024anysat}, a geospatial foundational model pre-trained on several remote sensing modalities, including the ones covered by some benchmarks considered in this study\footnote{The AnySat model is fine-tuned in each single-modality inference case.}, \ntext{FModDrop }\cite{chen2024novel}\ntext{ a multi-modal model using modality dropout over transformer layers, FCoM-av }\cite{mena2025maug},\ntext{ a multi-modal model simulating all combinations of missing modalities at training}, and  ESensI \cite{mena2024maclean}, an ensemble with shared prediction heads in the single-modality models.

\subsection{Experimental setting} \label{sec:setting}
We train all competitors having access to all multi-modal data during training, while the inference evaluation is performed considering a single modality, considering the all-but-one missing modality scenario; see Fig.~\ref{fig:colearn} for an illustration.
To measure performance, we use the \gls{f1} score for classification tasks and the \gls{r2} for the regression task. For the \gls{treesat} dataset, we use the validation set to select models and the test set for final evaluation.  As \gls{cropB}, \gls{cropM}, and \gls{lfmc} datasets do not have a predefined test partition available, we use a standard 10-fold cross-validation for evaluation, following common practices in the literature~\cite{rao2020sar,mena2025adaptive}. For each method, we report results averaged over 5 runs.

All the input data is rescaled via z-score normalization.
\ntext{
For the modality-specific encoders, we adopt standard architectures commonly used in the {\gls{eo}} literature. Specifically, for all multi-temporal modalities, we use TempCNN {\cite{pelletier2019temporal}}, a widely adopted backbone based on 1D convolutions over the temporal dimension. 
For the mono-temporal modality, i.e., the aerial image in {\gls{treesat}}, we use ResNet-50 {\cite{he2016deep}}, a 2D convolutional neural network with residual connections. 
In each encoder, an additional projection layer is used—a linear layer of 128 units with 20\% dropout. 
For parameters optimization, we use the Adam optimizer with a learning rate of {$10^{-3}$}, batch size of 128, and early stopping with a patience of 5, across 100 training epochs. For all competitors, we retain the default hyperparameter settings as reported in their original works.
}

\ntext{
In our framework, we intentionally opt for a simple and consistent design across modalities to ensure reproducibility and reduce model complexity. 
Thus, all prediction heads—i.e., $\letterfunc{P}_m(\cdot)$, $\letterfunc{P}^{\text{aux}}(\cdot)$, $\letterfunc{P}^{\text{spe}}(\cdot)$, and $\letterfunc{P}^{\text{un}}(\cdot)$ for both modalities $m \in \{1,2\}$—are implemented as single linear layers. We present results using alternative encoder backbones in Sec.~{\ref{sec:enco}}, showing that the proposed framework performs consistently across architectures. We did not perform an extensive hyperparameter search, as our architectural choices are grounded in standard practices from the literature. }

\ntext{
We use two criteria for balancing the predictive losses ($\mathcal{L}_{\text{main}}$ and $\mathcal{L}_{\text{aux}}$) in our framework.
In the classification tasks, we have used a weighted cross-entropy with per-class weights that are inversely proportional to the number of samples. This is used to handle class imbalanced scenarios. In the regression task, we have applied a z-score normalization to the target. This is used to rescale the squared error loss and prevent it from dominating over others, like $\mathcal{L}_{\text{mod}}$.
Moreover, to avoid overfitting to any specific configuration, we adopt a uniform sum for aggregating the loss terms (Eq.~{\ref{eq:loss}}). In Sec.~{\ref{sec:abla}}, we report a comparative analysis with an adaptive weighting scheme, which further supports the effectiveness of our design. }

\subsection{Results} \label{sec:results}

We report the predictive performance results in Table~\ref{tab:res:cropb}, \ref{tab:res:cropm}, \ref{tab:res:tree}, and \ref{tab:res:lfmc}, for datasets \gls{cropB}, \gls{cropM}, \gls{treesat}, and \gls{lfmc} respectively.

\begin{table}[t!]
	\centering
    	\caption{\gls{f1} scores in the \gls{cropB} dataset, cropland (binary) classification. The \best{best} and \secondbest{second-best} values are highlighted. $^{\star}$Based on pre-training.}  \label{tab:res:cropb}
	\begin{tabularx}{0.9\textwidth}{ll|CC} \toprule
       	Method 	& Field  	& {Sentinel-1}  & {Sentinel-2} \\
             	\midrule 
    	MMGF & Multi-modal fusion & \multicolumn{2}{c}{${82.3}$} \\
    	\midrule 
     	{Individual}
            	& Unimodal learning &  $71.5$ &  $81.9$ \\
            	\midrule
            	EmbraceNet & Fusion with missing modalities & $68.8$ & $81.4$ \\
	ShaSpec & Fusion with missing modalities & $68.9$ & $76.1$  \\
	DeCuR$^{\star}$ & Multi-modal co-learning & $71.5$ & $81.6$ \\
	DML & Multi-modal co-learning & $\secondbest{71.7}$ & $81.5$ \\
	DisCoM-KD & Cross-modal distillation & $70.1$ & $78.6$ \\
	AnySat$^{\star}$ & (\gls{eo}) Fusion with missing modalities & $70.2$ & $\secondbest{82.4}$\\
    FModDrop & (\gls{eo}) Fusion with missing modalities & $70.6$ & $79.8$ \\
    FCoM-av & (\gls{eo}) Fusion with missing modalities & $71.4$ & $82.1$  \\
    ESensI & (\gls{eo}) Multi-modal co-learning & $\secondbest{71.7}$ & $81.7$ \\
     \textbf{\gls{our}} & Multi-modal co-learning & $\best{73.5}$ &  $\best{83.3}$ \\
    	\bottomrule
	\end{tabularx}
\end{table}

\begin{table}[t!]
	\centering
    	\caption{\gls{f1} scores in the \gls{cropM} dataset, crop-type (multi-class) classification. The \best{best} and \secondbest{second-best} values are highlighted. $^{\star}$Based on pre-training.}  \label{tab:res:cropm}
	\begin{tabularx}{0.9\textwidth}{ll|CC} \toprule
       	Method 	& Field  	& {Sentinel-1} & {Sentinel-2} \\
             	\midrule 
    	MMGF & Multi-modal fusion & \multicolumn{2}{c}{$73.3$} \\
    	\midrule 
     	{Individual}
            	& Unimodal learning &  $55.4$ & $72.2$ \\
            	\midrule
            	EmbraceNet & Fusion with missing modalities & $40.6$ & $69.4$ \\
	ShaSpec & Fusion with missing modalities & $44.3$ & $63.9$ \\
	DeCuR$^{\star}$ & Multi-modal co-learning & $55.1$ & $71.9$ \\
	DML & Multi-modal co-learning & $55.1$ & $71.7$ \\
	DisCoM-KD & Cross-modal distillation & $\secondbest{55.6}$  & $70.2$\\
	AnySat$^{\star}$ & (\gls{eo}) Fusion with missing modalities & $52.2$ & $\secondbest{73.8}$ \\
    FModDrop  & (\gls{eo}) Fusion with missing modalities & $52.5$ & $69.7$ \\ 
    FCoM-av & (\gls{eo}) Fusion with missing modalities & $54.9$ & $72.3$ \\
	ESensI & (\gls{eo}) Multi-modal co-learning & $\secondbest{55.6}$ & $71.8$  \\
     \textbf{\gls{our}} & Multi-modal co-learning & $\best{58.3}$ & $\best{74.2}$ \\
    	\bottomrule
	\end{tabularx}
\end{table}

For the classification datasets (\gls{cropB}, \gls{cropM}, \gls{treesat}), we observe that methods proposed in the \gls{eo} domain, like AnySat, \ntext{FCoM-av,} and ESensI, tend to work better than approaches proposed in the more generic computer vision and machine learning domains. This is reasonable, as data in the \gls{eo} domain have a heterogeneous nature and require approaches that explicitly consider the \ntext{peculiar} characteristics of this kind of information \cite{rolf2024mission}. Concerning the rest of the competitors, we can note that ShaSpec, a state-of-the-art method for missing image modalities, clearly exhibits poor results on \gls{treesat} and \gls{lfmc}, as well as EmbraceNet in the \gls{lfmc} dataset. Moreover, existing literature has shown that methods introduced for data fusion in the context of missing modalities are not sufficiently robust and generic for single-modality predictions~\cite{mena2025maug}.
This fact underscores the need for specialized frameworks for multi-modal co-learning that can effectively handle both the complexity and heterogeneity of multi-modal data, as well as the missing modality \ntext{scenario at inference time.}

\begin{table}[t!]
	\centering
	\caption{\gls{f1} scores in the \gls{treesat} dataset, tree (multi-label) classification. The \best{best} and \secondbest{second-best} values are highlighted. $^{\star}$Based on pre-training. $\dagger$It only predicts a no-label pattern.}  \label{tab:res:tree}
	\begin{tabularx}{0.9\textwidth}{ll|CC} \toprule
    	Method    	&  Field  	& Aerial  & Sentinel-2 \\
            	\midrule 
    	OmniSAT  & Multi-modal fusion & \multicolumn{2}{c}{73.3} \\
    	MMGF & Multi-modal fusion & \multicolumn{2}{c}{68.6}  \\
    	\midrule 
     	Individual
            	& Unimodal learning &  $64.7$ & $64.9$ \\
            	\midrule
	EmbraceNet & Fusion with missing modalities & $63.2$ & $47.3$ \\
	ShaSpec & Fusion with missing modalities & $\dagger$ & $\dagger$ \\
    	DeCuR$^{\star}$ & Multi-modal co-learning & $62.4$ & $64.2$ \\
	DML & Multi-modal co-learning & ${65.2}$ & $60.5$ \\
	DisCoM-KD & Cross-modal distillation & $60.4$ & $47.9$ \\
	AnySat$^{\star}$ & (\gls{eo}) Fusion with missing modalities & $60.0$ & $\best{74.8}$ \\
    FModDrop  & (\gls{eo}) Fusion with missing modalities & $59.1$ & $54.0$  \\
    FCoM-av & (\gls{eo}) Fusion with missing modalities & $\secondbest{65.9}$ & $60.2$ \\
	ESensI & (\gls{eo}) Multi-modal co-learning & ${64.3}$ & $57.9$ \\
    \textbf{\gls{our}} & Multi-modal co-learning & $\best{66.4}$  & $\secondbest{66.3}$   \\
    	\bottomrule
	\end{tabularx}
\end{table}

\begin{table}[t!]
	\centering
	\caption{\gls{r2} scores in the \gls{lfmc} dataset, {regression}. The \best{best} and \secondbest{second-best} values are highlighted. $^{\star}$Based on pre-training. }  \label{tab:res:lfmc}
	\begin{tabularx}{0.9\textwidth}{ll|CC} \toprule
       	Method 	& Field  	& {Sentinel-1} & {Landsat-8} \\
             	\midrule 
    	InputFu & Multi-modal fusion & \multicolumn{2}{c}{$0.520$}  \\
    	\midrule 
     	{Individual}  
            	& Unimodal learning  &  $0.242$ & $0.432$ \\
    	\midrule
    	EmbraceNet & Fusion with missing modalities & $0.110$ & $0.079$ \\
	ShaSpec & Fusion with missing modalities & $0.053$ & $0.189$ \\
	DeCuR$^{\star}$ & Multi-modal co-learning & $\best{0.248}$ & $\secondbest{0.417}$ \\
	DisCoM-KD & Cross-modal distillation & $0.226$ & $0.416$ \\
	AnySat$^{\star}$ & (\gls{eo}) Fusion with missing modalities & $0.185$ & $0.353$ \\
    FModDrop  & (\gls{eo}) Fusion with missing modalities & $0.080$ & $0.280$ \\
    FCoM-av  & (\gls{eo}) Fusion with missing modalities & $0.214$ & $0.374$ \\
	ESensI & (\gls{eo}) Multi-modal co-learning & $\secondbest{0.233}$ & $0.393$ \\
    \textbf{\gls{our}}  & Multi-modal co-learning & $\best{0.248}$ & $\best{0.467}$   \\
	\bottomrule
	\end{tabularx}
\end{table}

Throughout the benchmarks, our \gls{our} framework consistently outperforms both Individual baselines and all competing methods in nearly all considered cases.
The improvement over the single-modality baseline (Individual) is not straightforward, as several competitors fail to \ntext{ameliorate} these baselines. For example, this is the case for DisCoM-KD on \gls{cropB} and EmbraceNet on \gls{cropM}.
\ntext{Our approach achieves comparable performances to DeCuR in the Sentinel-1 evaluation of the LFMC dataset}, and is only outperformed by AnySat in the Sentinel-2 evaluation of the \gls{treesat} dataset, which can be attributed to AnySat's pre-training exposure to this specific benchmark data. Despite these isolated cases, our framework achieves superior results in the other inference modality (Aerial).
Furthermore, our performance improvement is substantial in several cases, particularly with both modalities in \gls{cropM} and with Landsat-8 modality in the \gls{lfmc} dataset. Regarding the CropHarvest benchmarks (\gls{cropB} and \gls{cropM}), our approach even improves the results of the multi-modal reference, MMGF, that leverages both data modalities when the Sentinel-2 modality is considered for inference.

Overall, we notice that the methods obtaining the second-best results, across the benchmarks, vary depending on which modality is available for inference, while \gls{our} obtains consistent improvements no matter the considered modality.
This indicates that our framework takes advantage of modality cooperation in the multi-modal training stage, enhancing the single-modality model's performance at inference time. 

\subsection{Ablation}\label{sec:abla}

We conduct an ablation study to isolate the key factors characterizing the behavior of \gls{our}. To this end, we individually remove components in our framework, considering the \gls{cropB} and \gls{cropM} datasets. These results are reported in Table~\ref{tab:abl:cropb}.

\begin{table}[t!]
	\centering
    	\caption{\gls{f1} scores in the binary and multi-class classification tasks for different variations in our framework. The \best{best} and \secondbest{second-best} values are highlighted.}  \label{tab:abl:cropb}
	\begin{tabularx}{0.9\textwidth}{l|cc|cc|C} \toprule
 	& \multicolumn{2}{c|}{\gls{cropB}} & \multicolumn{2}{c|}{\gls{cropM}} \\ \cmidrule{2-6}
       	Variant	& {Sentinel-1}  & {Sentinel-2} &  {Sentinel-1}  & {Sentinel-2} & Avg. \\
             	\midrule 
     	{Individual}  &  $71.5$ &  $81.9$ &  $55.4$ & $72.2$ & $70.3$ \\
            	\midrule
        \textbf{\gls{our}} & $\best{73.5}$ & $\best{83.3}$ & $\best{58.3}$ & $\best{74.2}$ & $\best{72.3}$ \\ \cmidrule{2-6}
        $\rightarrow$ w/o modality loss & $\secondbest{73.3}$ &  $\secondbest{83.0}$ & $\secondbest{58.1}$ & $\secondbest{74.0}$ & $\secondbest{72.1}$ \\
        $\rightarrow$ w/o auxiliary loss & $73.1$ & $82.9$ & $58.0$ & $\best{74.2}$ & $72.0$ \\
        $\rightarrow$ w/o contrastive loss & $70.4$ & $80.2$  & $56.5$  & $71.6$ & $69.7$ \\
        $\rightarrow$ w/o unused features  & $72.8$ & $82.8$  & $56.5$ & $72.7$ & $71.2$ \\
     	$\rightarrow$ unpaired data & $71.8$ & $82.1$ & $57.4$ & $\secondbest{74.0}$ & $71.3$  \\
     	$\rightarrow$ shared encoders & $72.8$ & $82.8$ & $56.0$ & $72.4$ & $71.0$ \\
        $\rightarrow$ weighted loss & $72.4$ & $82.7$ & $57.8$ & $\best{74.2}$ & $71.2$ \\
    	\bottomrule
	\end{tabularx}
\end{table}

We note that the contrastive loss has the greatest impact on model performance when removed, while the \ntext{modality-discriminant and auxiliary predictive losses} demonstrate less individual impact. 
\ntext{One possible explanation is that the contrastive loss is greatly enhancing the cross-modal interaction by aligning the representation of heterogeneous sensor modalities. In comparison, the other loss terms may regulate modality-specific representations in a more localized manner. We also note that the contrastive loss typically holds higher values during training (Sec.~{\ref{sec:losses}}), suggesting its dominant role in driving the co-learning process. }
Nevertheless, the combination of all the loss terms enables our approach to fully exploit the available multi-modal data during training and clearly improves single-modality inference overall. 

Furthermore, we consider an additional ablation analysis of our framework where we discard the unused feature component (Sec.~\ref{sec:method:spaces}). This experiment leads to a drop of around 1 point in the F1 score. The same relative drop is observed when experimenting using unpaired modalities.
Moreover, we consider an ablation case where the same encoder is used to extract the per-modality unique and common information (i.e. $\letterfunc{E}_m^{\text{com}} = \letterfunc{E}_m^{\text{uni}}$ in Sec.~\ref{sec:method:spaces}). This variation produces a considerable decrease in performance in the \gls{cropM} dataset of around 2 points in the F1 score.

\ntext{Additionally, we include a comparison to a version of our framework including an adaptive weighting strategy to combine all loss terms (Eq.~{\ref{eq:loss}}). More precisely, we follow the uncertainty-based criteria by Kendall et al. {\cite{kendall2018multi}}. Notably, this strategy yields inferior results compared to the simple summation we have adopted. This reflects that our framework benefits from an unbiased aggregation of the loss terms to improve the co-learning process and thereby, single-modality inference. 
}

The ablation analysis indicates that all the components on which our \gls{our} framework is built, and their interplay, contribute to its superior behavior in the  (all-but-one) missing modality scenario. Moreover, this occurs regardless of the considered benchmark and available modality at the inference stage.

\subsection{Usefulness of shared and specific features} \label{sec:shaspe}

\begin{table}[t!]
	\centering
    	\caption{Predictive performance when using only the shared or specific features in the \gls{our} framework. The \best{best} and \secondbest{second-best} values between our variants are highlighted.}  \label{tab:abl:featurespace}
	\begin{tabularx}{\textwidth}{l|CC|CC|CC|CC} \toprule
 	& \multicolumn{2}{c|}{\gls{cropB}} & \multicolumn{2}{c|}{\gls{cropM}} & \multicolumn{2}{c|}{\gls{treesat}} & \multicolumn{2}{c}{\gls{lfmc}} \\ \cmidrule{2-9}
       	Variant	& {S1}  & {S2} &  {S1}  & {S2} & Aerial & S2 & S1 & L8\\
             	\midrule 
            	\midrule
        \textbf{\gls{our}} & $\best{73.5}$ &  $\best{83.3}$ & $\best{58.3}$ & $\best{74.2}$ &  $\best{66.4}$ & $\best{66.3}$ & $\best{0.248}$ & $\best{0.467}$ \\ \cmidrule{2-9}
        $\rightarrow$ shared features & $\secondbest{71.7}$ & $81.6$ & $\best{58.3}$ & $70.5$ & $35.3$ & $28.4$ & $0.226$ & $\secondbest{0.445}$ \\
        $\rightarrow$  specific features & $71.5$ & $\secondbest{82.3}$  & $\secondbest{55.3}$ & $\secondbest{72.8}$ & $\secondbest{65.3}$ & $\secondbest{63.5}$ & $\secondbest{0.230}$ & $0.421$  \\
    	\bottomrule
	\end{tabularx}
\end{table}
Table~\ref{tab:abl:featurespace} shows the performance when using only the shared or specific features learned by \gls{our}. Overall, the method achieves the best results when both feature sets are considered, highlighting the benefit of leveraging complementary information. In most cases, specific features contribute more positively to performance than shared ones. An exception is observed in the Sentinel-1 modality of \gls{cropM}, where shared features slightly outperform specific ones. 
Interestingly, results using only shared features are notably lower for the \gls{treesat} dataset. 
This may be due to the challenge of aligning multi-modal data captured by sensors with substantial differences in spatial and temporal resolutions, such as Sentinel-2 and Aerial. In contrast, sensor pairs with more similar characteristics (e.g., Sentinel-1 and Sentinel-2) tend to encode more \ntext{common} information relevant to the task.  \ntext{These observations indicate that the relationship between modality-shared and modality-specific features is both sensor- and task-dependent, with no absolute rule about the balance between these two spaces.}

\subsection{Effect of different encoders} \label{sec:enco}

To analyze the sensitivity of our \gls{our} framework to the modality encoders, we evaluate its performance using various backbones in the multi-temporal modalities.  The goal is to verify whether the learning process remains consistent across architectures.  To this end, we consider several temporal encoders with diverse characteristics in addition to the TempCNN backbone. These are: i) a two layer Multi-Layer Perceptron (MLP) that \ntext{ignores} the temporal dimension in the input modalities, ii) a standard Transformer \cite{vaswani2017attention} that explicitly models temporal dependencies, iii) ConvTran \cite{foumani2024improving}, a recent backbone for \ntext{multi-variate time series classification} that combines both 1D convolutions with transformer blocks, and iv) an LSTM \cite{hochreiter1997long} (Long Short Term Memory) recurrent neural network.
The results on the \gls{lfmc} dataset are shown in Fig.~\ref{fig:res:encoders}, while others are in the appendix (Sec.~\ref{appendix:enc}). 
\begin{figure}[!t]
	\centering
	\includegraphics[width=\textwidth]{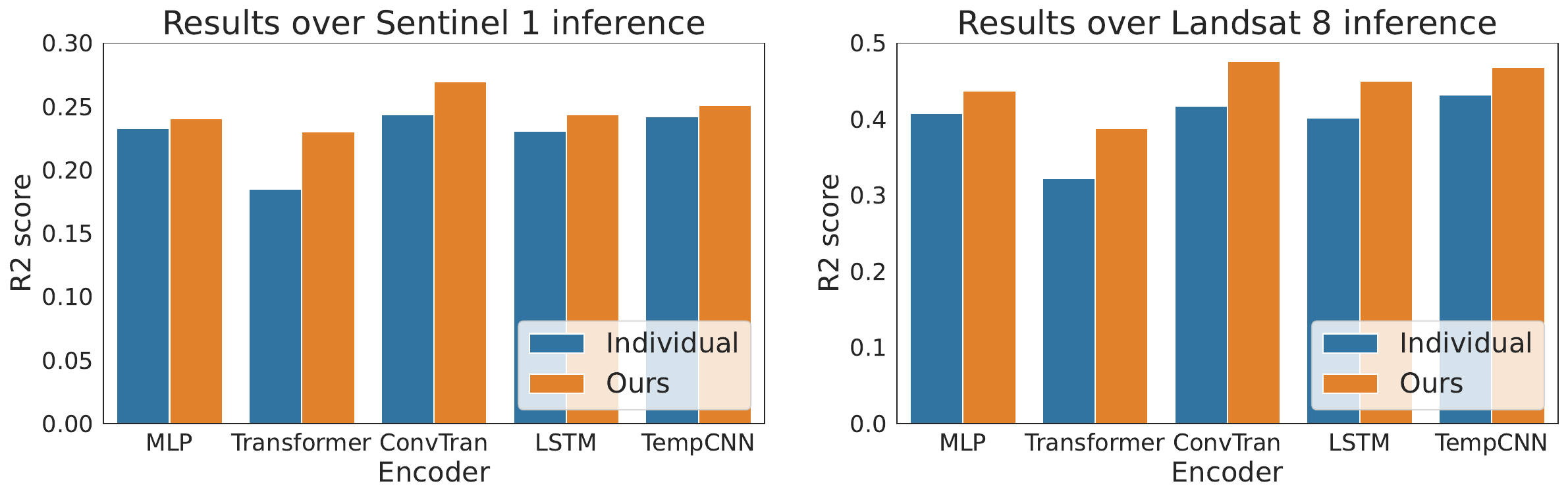}
	\caption{Predictive performance by using different encoder architectures in the \gls{our} framework. Results are from the \gls{lfmc} dataset (regression).}\label{fig:res:encoders}
\end{figure}

We observe that the performance of our framework can be influenced by the encoder architecture, with the best results coming from the ConvTran, TempCNN, and LSTM encoders, respectively. 
\ntext{Surprisingly, the worst results are obtained with the Transformer architecture. Although its layer structure is comparable to other models (see Sec.~{\ref{appendix:enc}}), this may be attributed to the over-parametrization of this model--over one million parameters (Table~{\ref{tab_app:parameters}})--which can hamper the learning process from the relatively small {\gls{lfmc}} dataset (fewer than 3 thousand samples). Notably, the Transformer and ConvTran architectures achieve the top performances on the other datasets (see Sec.~{\ref{appendix:enc}} in the appendix). 
}
Moreover, we notice that the improvement of \gls{our}, compared to the individual trained models, is consistent across different encoder architectures. This relative gain is higher when recent backbone encoders (i.e., ConvTran) are employed, while it is moderate with TempCNN and MLP. Overall, this analysis demonstrates the general applicability of our framework regardless of the considered encoder architecture.

\subsection{Qualitative analysis of learned features} \label{sec:quali}

\begin{figure}[!t]
	\centering
	\includegraphics[width=\textwidth]{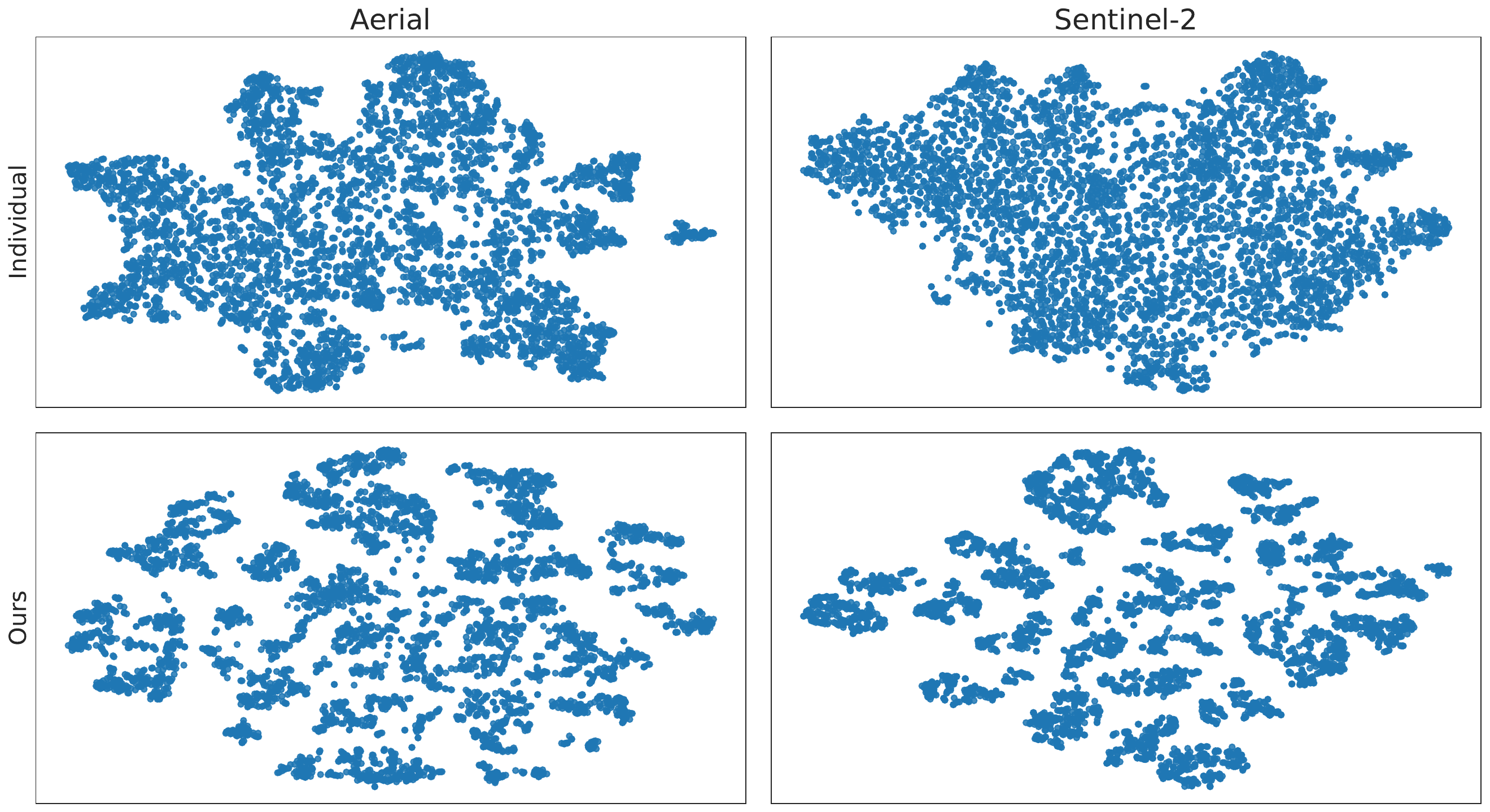}
	\caption{t-SNE projection of the learned features by the Individual method and our \gls{our} (concatenation of shared and specific features) on the \gls{treesat} dataset.}\label{fig:res:tsne}
\end{figure}
We visually inspect the learned features using our \gls{our} framework in a 2D projection via the t-SNE method \cite{van2008visualizing}, illustrated in Fig.~\ref{fig:res:tsne}.
We compare the features extracted by our framework (concatenation of shared and specific ones) with the ones derived by the Individual methods in the multi-label dataset \gls{treesat}. Thus, we observe that the features learned by \gls{our} exhibit a clear cluster structure compared to the features of the Individual single-modality models. This pattern is observed in the features learned for both Aerial and Sentinel-2 modalities.

\subsection{Loss functions during training} \label{sec:losses}

\begin{figure}[!t]
	\centering
    {\includegraphics[width=0.48\textwidth, page=1]{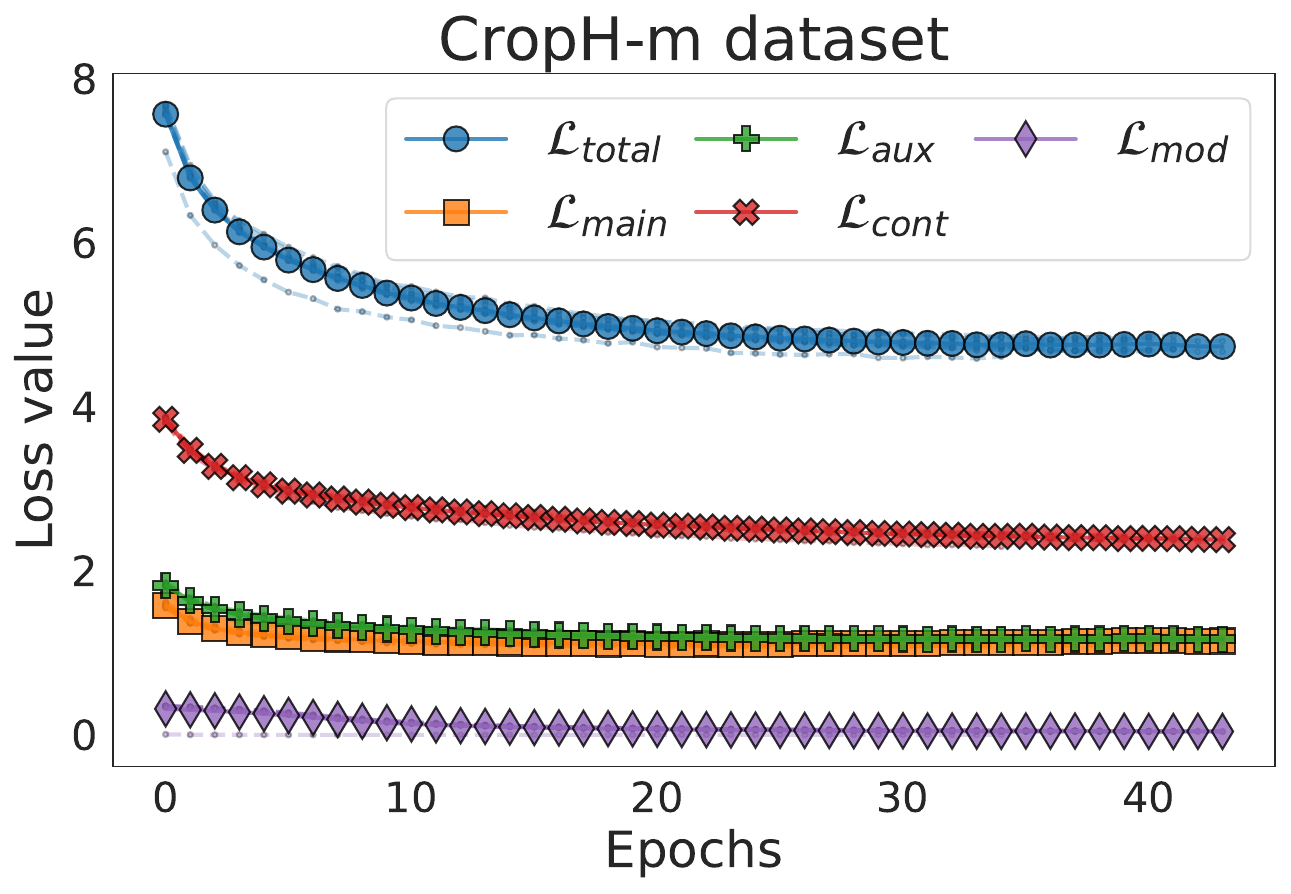}}
    \hfill
    {\includegraphics[width=0.48\textwidth, page=1]{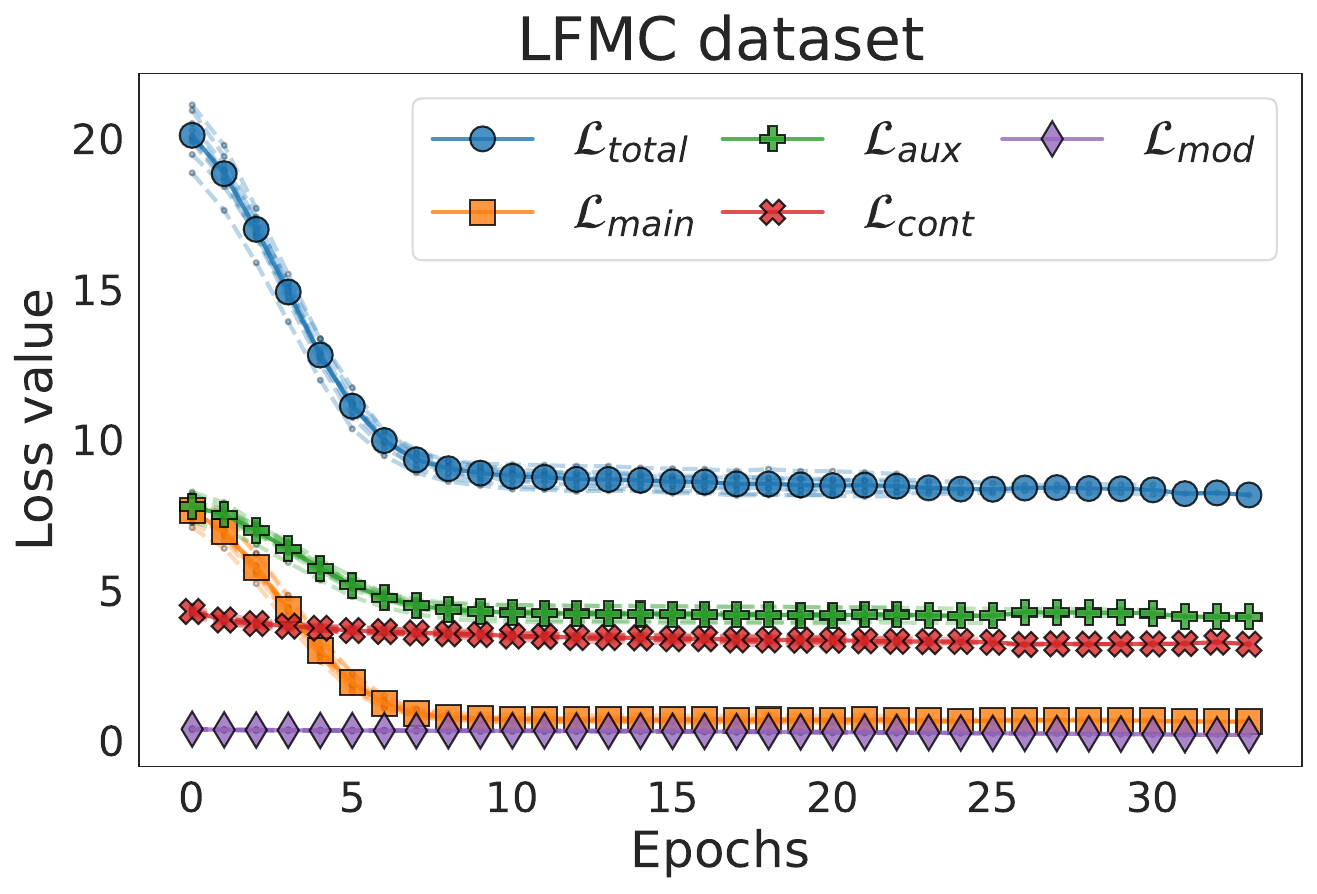}}
    \hfill
	\caption{\ntext{Individual loss functions of our framework across the training. The average across folds and multiple runs is shown for each loss function.}}\label{fig:losses}
\end{figure}
\ntext{
To analyze the relative magnitudes of the different loss functions employed in our framework, we plot their trends during training in the CropH-m and {\gls{lfmc}} datasets in Fig.~{\ref{fig:losses}}. The results of the CropH-b dataset are shown in the appendix (Sec.~{\ref{appendix:loss}}).
Under the classification task, all loss functions exhibit similar behaviors. For example, the modality discriminant minimization reaches the lowest magnitudes, whereas the contrastive loss maintains the highest ones.
In the regression task, the main predictive loss exhibits rapid convergence during the initial training epochs, whereas the contrastive and auxiliary losses maintain relatively high values throughout the training process. 
This behavior demonstrates that our framework achieves balanced optimization across all loss components, effectively preventing any single loss function from dominating the learning process.}

\section{Conclusion} \label{sec:conclu}

The possible lack of sensor modalities at inference time limits the applicability of multi-modal models in the \gls{eo} domain. In this work, we demonstrate that multi-modal co-learning, \ntext{ by allowing interaction between available modalities at training time, can enhance single-modality model performance at inference}. 
To achieve this, we introduce a novel task-agnostic framework \ntext{that combines multiple loss functions to enable effective feature-based collaboration} among heterogeneous sensor modalities.
Our approach \ntext{aims to disentangle modality}-shared, \ntext{modality}-specific, and unused information, thereby improving the performance of single-modality models.
We validate our framework on several \gls{eo} multi-modal benchmarks spanning binary, multi-class, multi-label classification, and regression tasks.
Our framework, unlike recent methods from the literature, systematically improves upon models trained on individual modalities and consistently outperforms state-of-the-art methods from the general fields of computer vision and machine learning, as well as \gls{eo}-specific strategies, across various downstream tasks.
Our research contribution to multi-modal co-learning advances the state of the art in the \gls{eo} domain under missing modality scenarios, precisely when only a single training modality remains accessible during inference.

\textbf{\emph{Limitations}.}\hspace{0.15cm}
The following points discuss the limitations of our work and suggest potential future research directions: 
i) The current framework has been validated for benchmarks with only two modalities available at the training stage. A natural extension would be to scenarios involving multiple modalities during training.
ii) Our approach has been validated exclusively on \gls{eo} data. 
Future work should extend the evaluation to general multi-modal computer vision and machine learning benchmarks.

\section*{Statements and Declarations}

\subsection*{Funding}
Francisco Mena acknowledges support through a scholarship of the University of Kaiserslautern-Landau.

\subsection*{Competing interests}
The authors declare that they have no conflicts of interest to report regarding the present study.

\bibliographystyle{ieeetr}
\bibliography{main}

\setcounter{table}{0}
\renewcommand{\thetable}{A\arabic{table}}
\setcounter{figure}{0}
\renewcommand{\thefigure}{A\arabic{figure}}
\appendix
\clearpage
\section{{Further Results}} \label{appendix}

\subsection{Encoders} \label{appendix:enc}

\begin{figure}[!h]
	\centering
	\includegraphics[width=\textwidth]{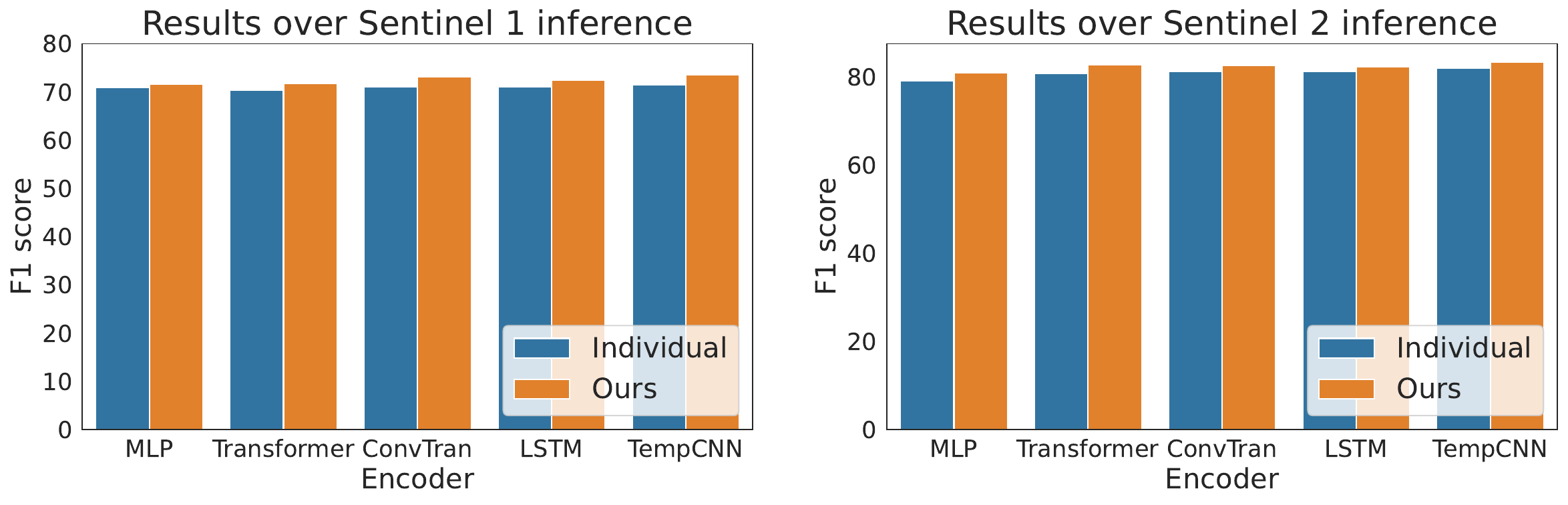}
	\caption{Predictive performance of the \gls{our} framework by using different encoder architectures in the \gls{cropB} dataset.}\label{fig_app:res:encoders:cropb}
\end{figure}

\begin{figure}[!t]
	\centering
	\includegraphics[width=\textwidth]{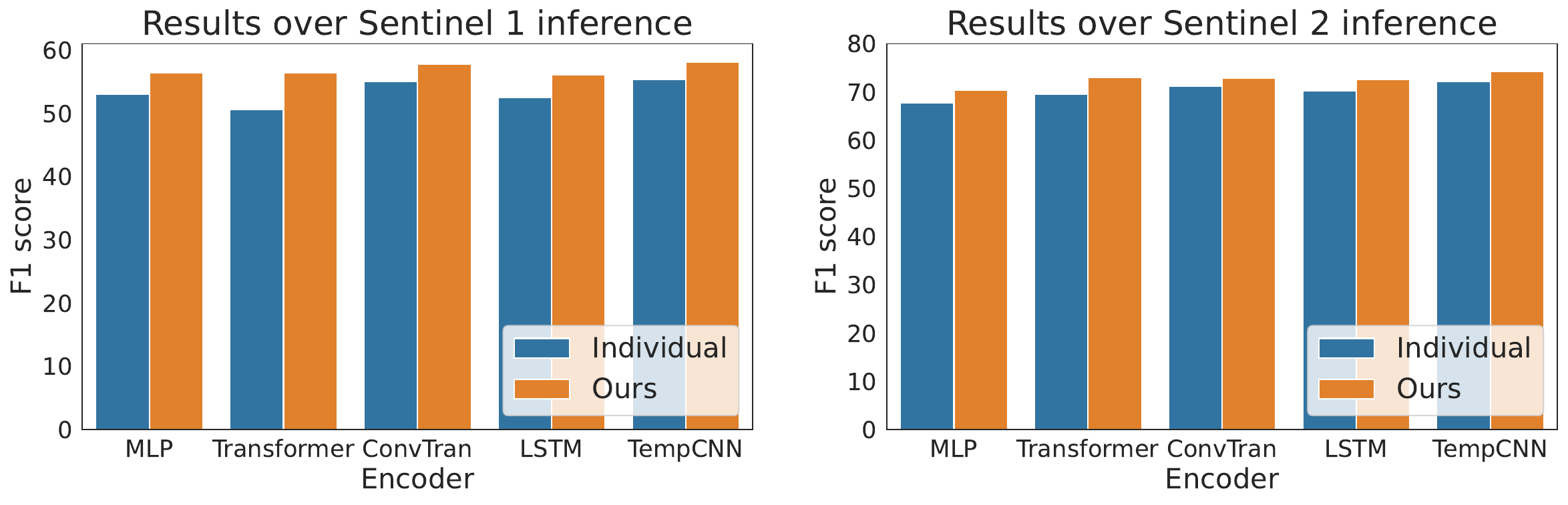}
	\caption{Predictive performance of the \gls{our} framework by using different encoder architectures in the \gls{cropM} dataset.}\label{fig_app:res:encoders:cropm}
\end{figure}

We replicate the analysis of our \gls{our} framework with different encoder architectures, made in Sec.~\ref{sec:enco}, but for the other datasets with multi-temporal modalities.
In addition, we display the number of learnable parameters in each of the encoders used along the datasets and modalities in Table~\ref{tab_app:parameters}.
We use 2 layers with 128 units in all encoders, with a few variants in the specific parameters. In concrete, we use a kernel size of 5 in the TempCNN, batch normalization layers in the MLP, and 8 head attentions in the Transformer and ConvTran.
Thus, Fig.~\ref{fig_app:res:encoders:cropb} and Fig.~\ref{fig_app:res:encoders:cropm} displays the results for the \gls{cropB} and \gls{cropM} datasets, respectivelly.
We observe a more stable performance of our framework across encoder architectures, compared to the ones in the \gls{lfmc} dataset (Fig.~\ref{fig:res:encoders}).
In all cases, the \gls{our} consistently outperforms the results of the individual trained models. 
Moreover, the best results of \gls{our} are obtained between TempCNN and ConvTran architectures.

\begin{table}[!h]
    \centering
    \caption{Number of learnable parameters along different encoder architectures tested. The values are in thousands.}
    \begin{tabular}{l|rr|rr|rr}  \hline
            & \multicolumn{2}{c|}{\gls{cropB} \& \gls{cropM} } & \multicolumn{2}{c|}{\gls{lfmc}} & \multicolumn{2}{c}{\gls{treesat}} \\ 
         Encoder & Sentinel 1 & Sentinel 2 & Sentinel 1 & Landsat 8 & Aerial & Sentinel 2 \\ \hline
         ResNet-50 \cite{he2016deep} && & &  & 23\,773  & \\
         TempCNN \cite{pelletier2019temporal} & 1\,052 & 1\,059 & 496 & 496 & & 10\,077 \\
         MLP & 36 & 52 & 35 & 37 &  \\
         Transformer \cite{vaswani2017attention} & 1\,203 & 1\,204 & 1\,203 & 1\,204 &  \\
         ConvTran \cite{foumani2024improving} & 335 & 991 & 399 & 727 & \\
         LSTM \cite{hochreiter1997long} & 216 & 221 & 217 & 219 & \\
         \hline
    \end{tabular}
    \label{tab_app:parameters}
\end{table}

\subsection{Visualization of learned features} \label{appendix:tsne}

\begin{figure}[!h]
	\centering
	\includegraphics[width=\textwidth]{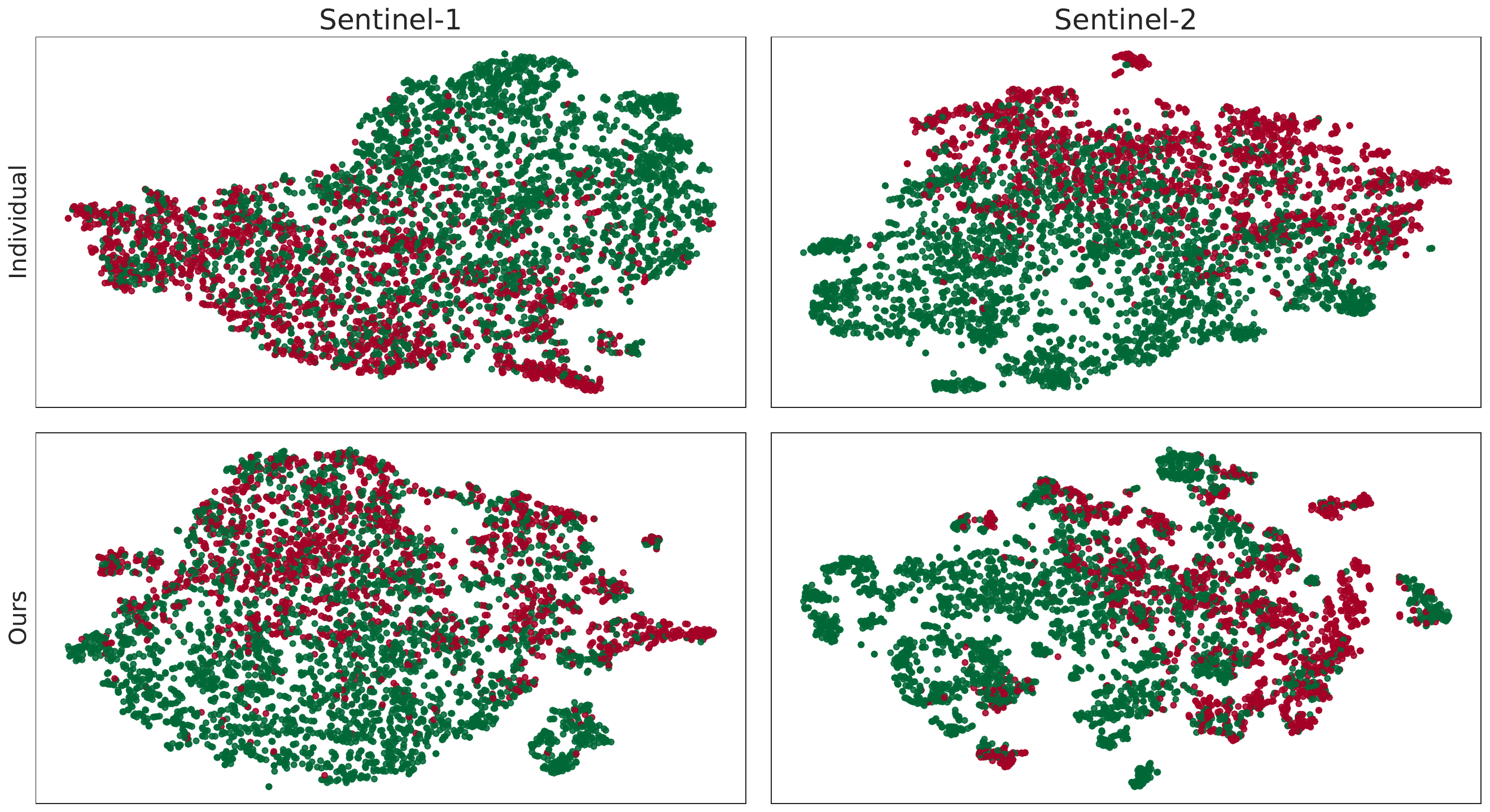}
    \caption{t-SNE projection of the learned features by the Individual method and our \gls{our} (concatenation of shared and specific features) on the \gls{cropB} dataset.}\label{fig_app:res:tsne}
\end{figure}

\begin{figure}[!h]
	\centering
	\includegraphics[width=\textwidth]{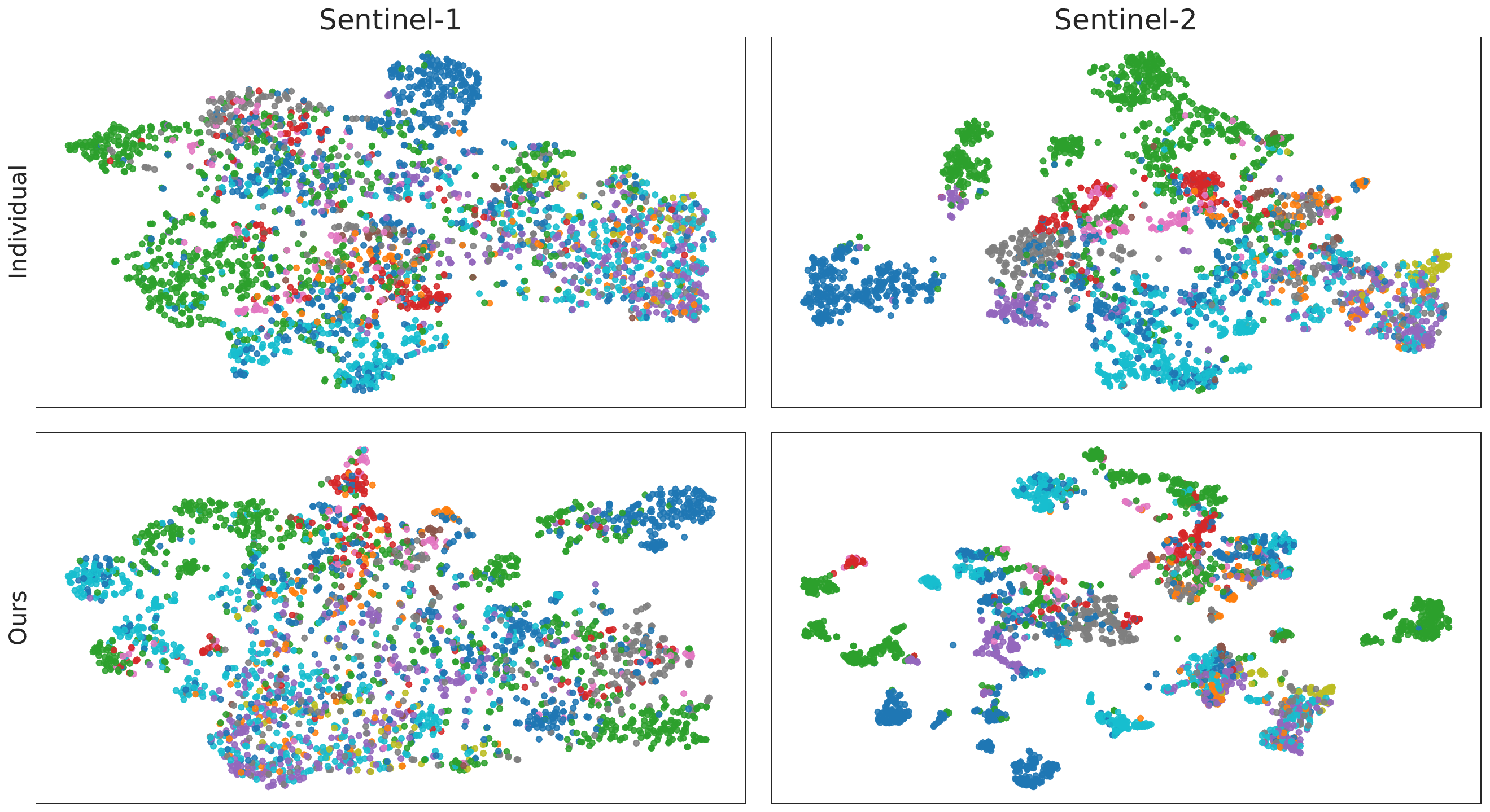}
	\caption{t-SNE projection of the learned features by the Individual method and our \gls{our} (concatenation of shared and specific features) on the \gls{cropM} dataset.}\label{fig_app:res:tsne2}
\end{figure}

We display the 2D projection of the learned features of our \gls{our} framework for the \gls{cropB} dataset in Fig.~\ref{fig_app:res:tsne} and the \gls{cropM} dataset in Fig.~\ref{fig_app:res:tsne2}.
The same findings are obtained from this analysis compared to the \gls{lfmc} dataset (in Fig.~\ref{fig:res:tsne}).
Moreover, we notice that the features learned by \gls{our} are grouped into subclusters of smaller size.


\subsection{Loss functions} \label{appendix:loss}

\begin{figure}[!h]
	\centering
    {\includegraphics[width=0.5\textwidth, page=1]{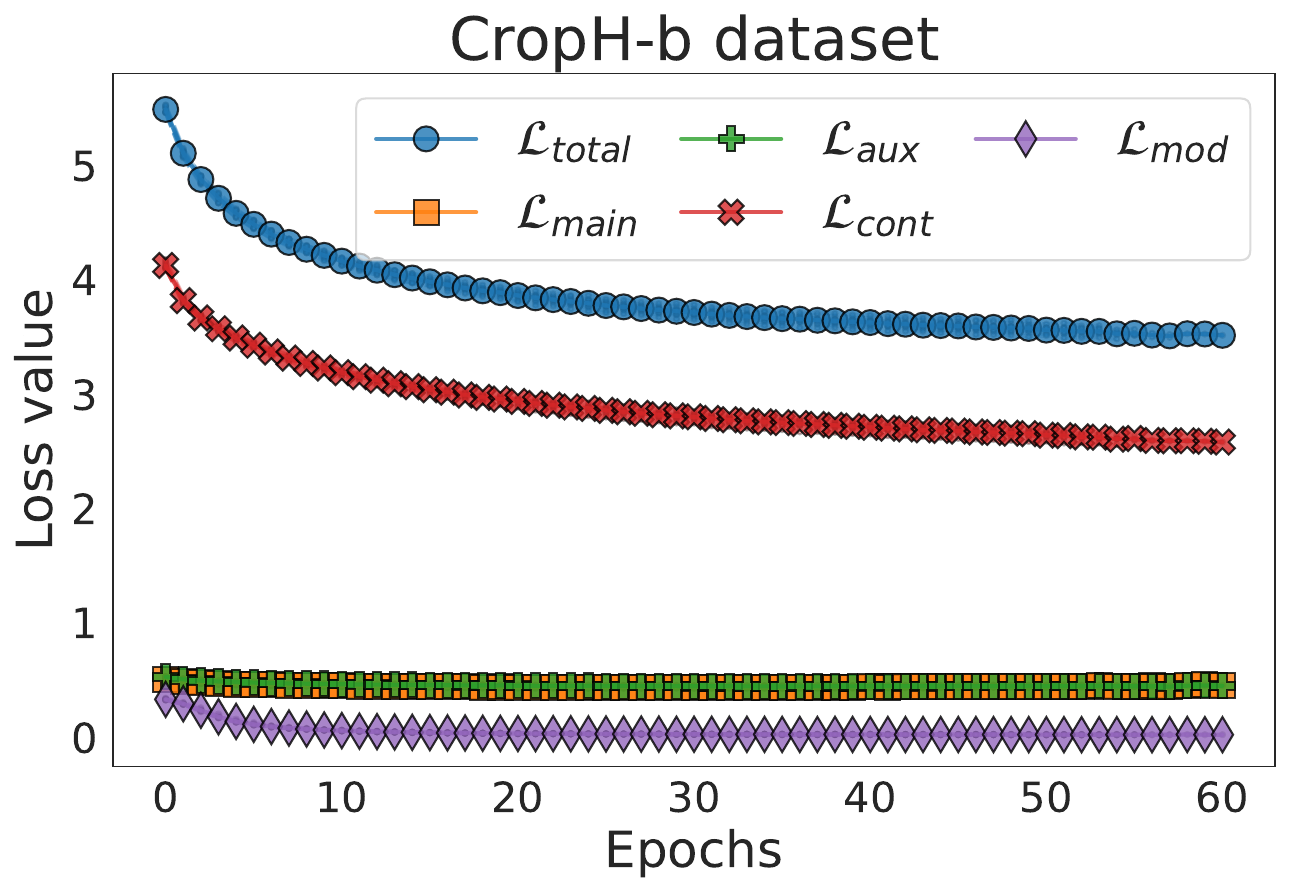}}
    \caption{Individual loss functions of our framework across the training. The average across folds and multiple runs is shown for each loss function.}\label{fig_app:losses}
\end{figure}

We show the relative magnitudes of the different loss functions employed in our framework in the \gls{cropB} dataset in Fig.~{\ref{fig_app:losses}}. 
We notice the same behavior described in Sec.~\ref{sec:losses} for the classification case, i.e. all loss functions exhibit similar minimization trends. 
Besides, the contrastive loss maintains the higher magnitudes, while modality discriminant reaches the lowest ones.

\subsection{Case study} \label{appendix:case}

\begin{figure}[!b]
	\centering
    {\includegraphics[width=0.32\textwidth, page=1]{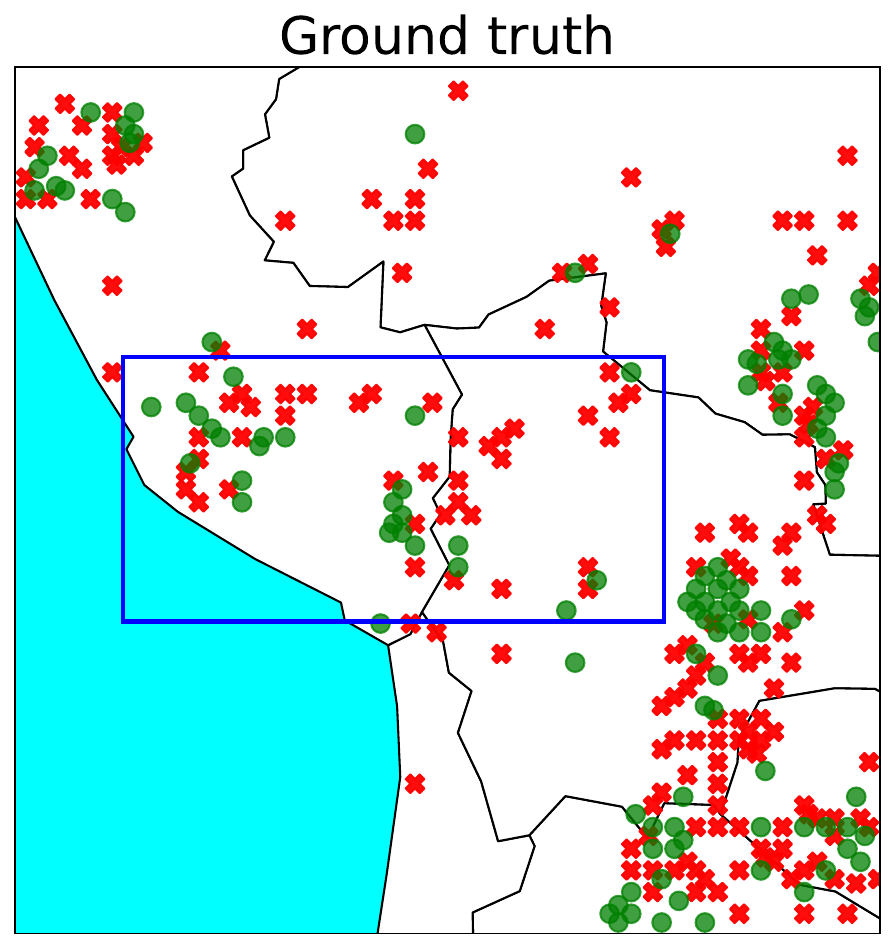}}
    \hfill
    {\includegraphics[width=0.32\textwidth, page=1]{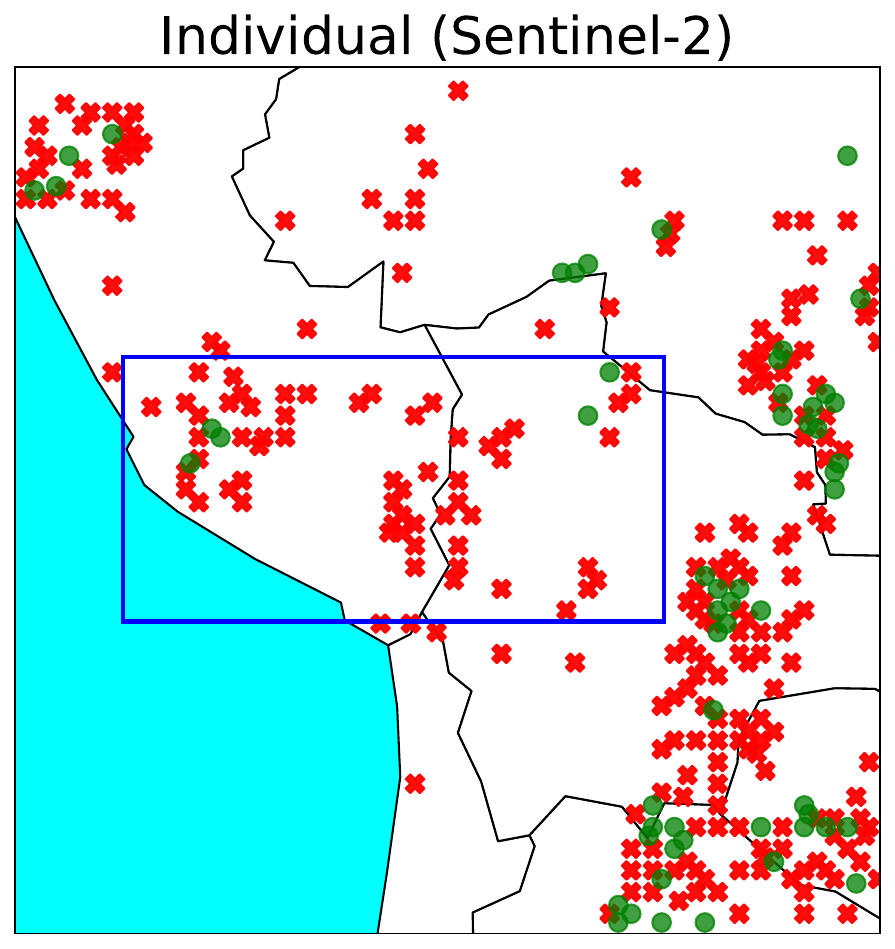}}
    \hfill
    {\includegraphics[width=0.32\textwidth, page=1]{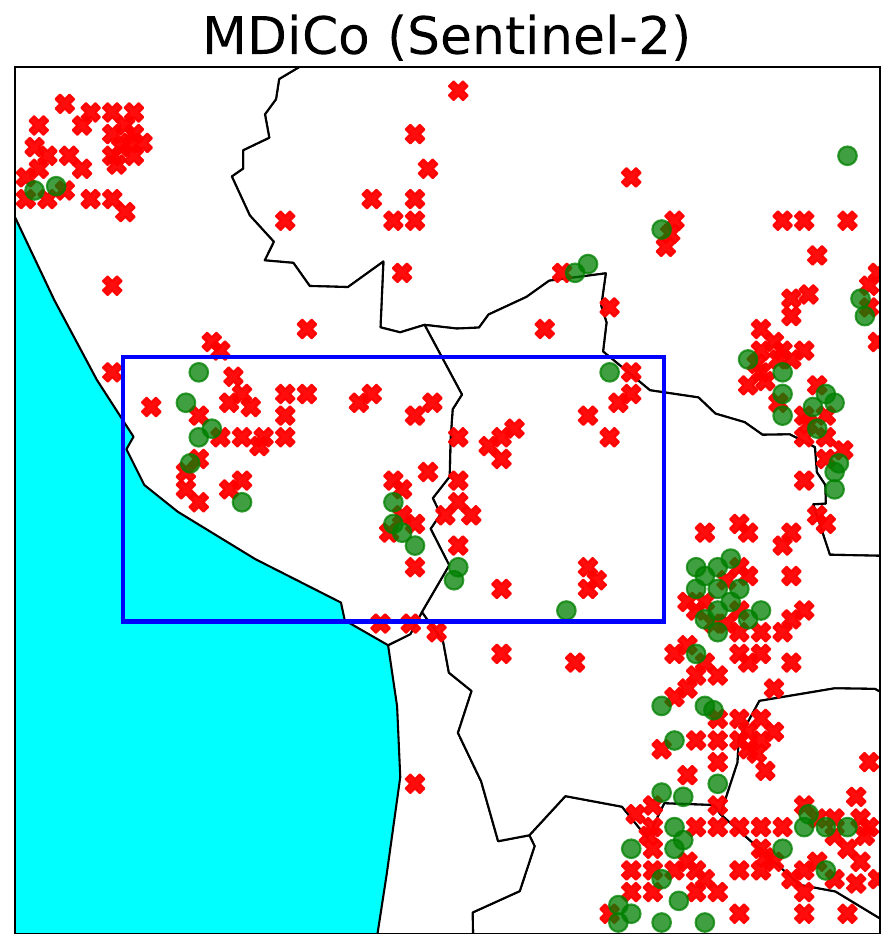}}
    \hfill
	\caption{Prediction map of our \gls{our} method in comparison to the ground truth labels and the Individual baseline using the Sentinel-2 modality. The region is in the Midwest of South America.}\label{appendix:fig:case1}
\end{figure}

\begin{figure}[!b]
	\centering
    {\includegraphics[width=0.32\textwidth, page=1]{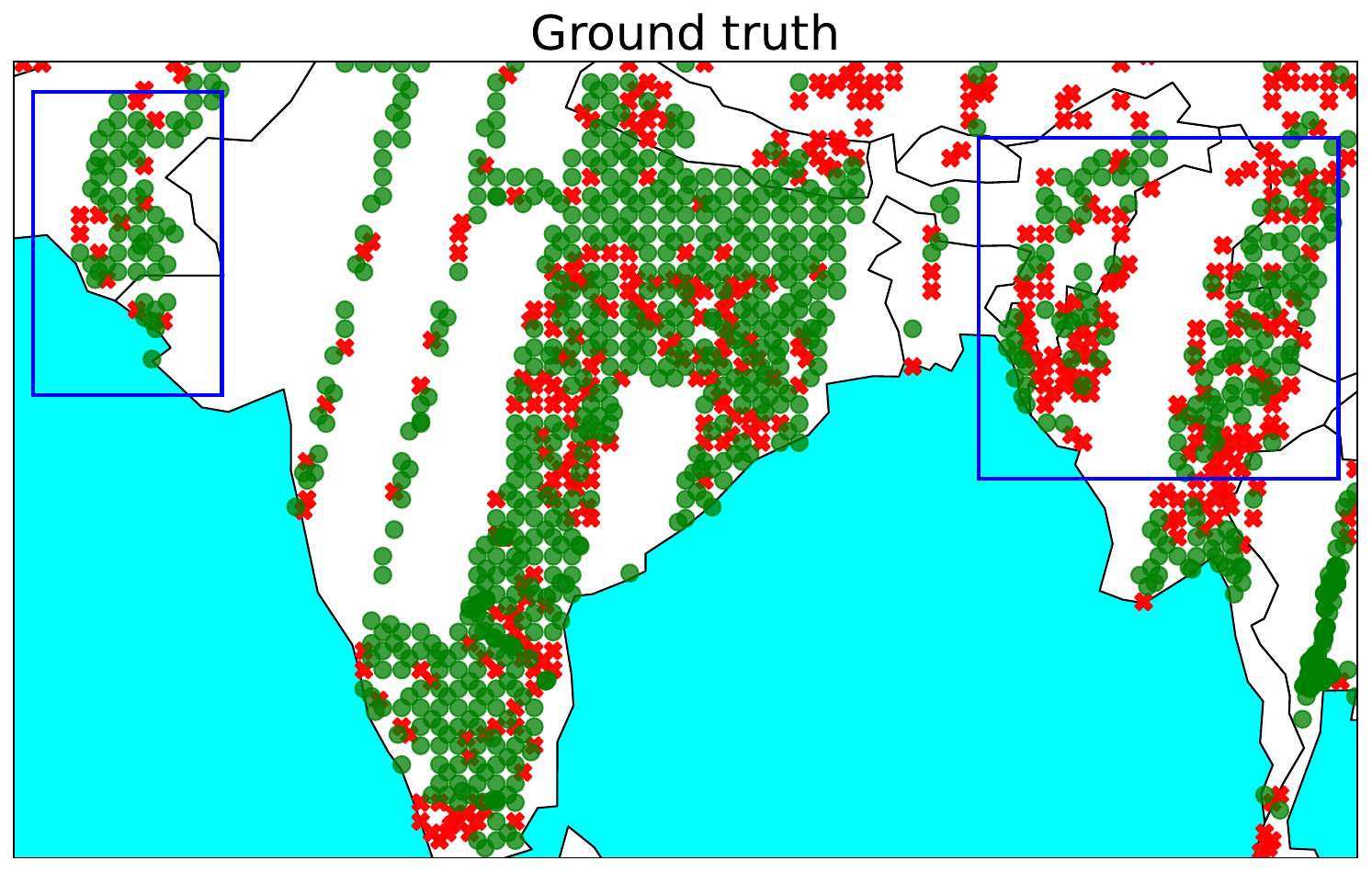}}
    \hfill
    {\includegraphics[width=0.32\textwidth, page=1]{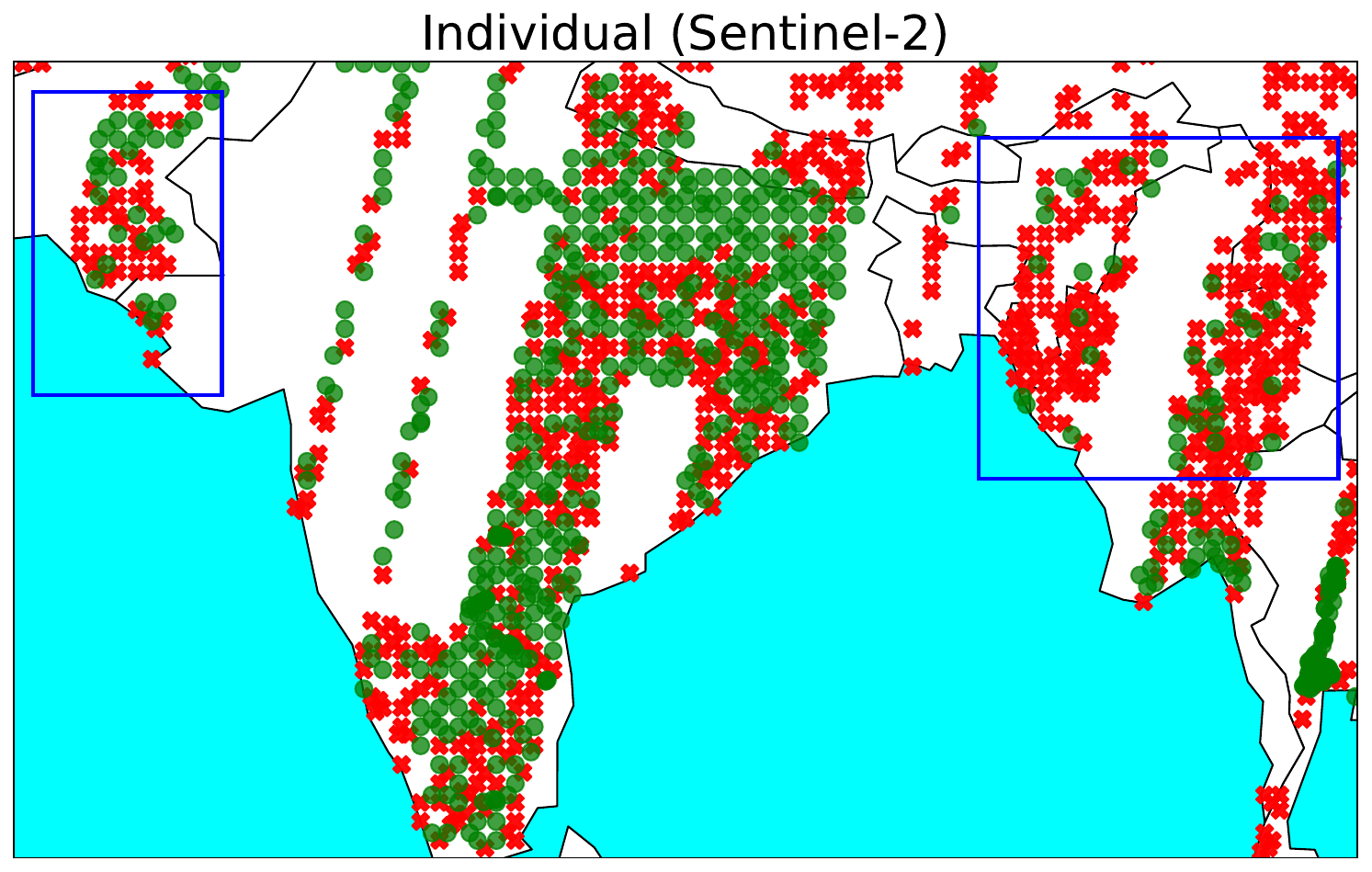}}
    \hfill
    {\includegraphics[width=0.32\textwidth, page=1]{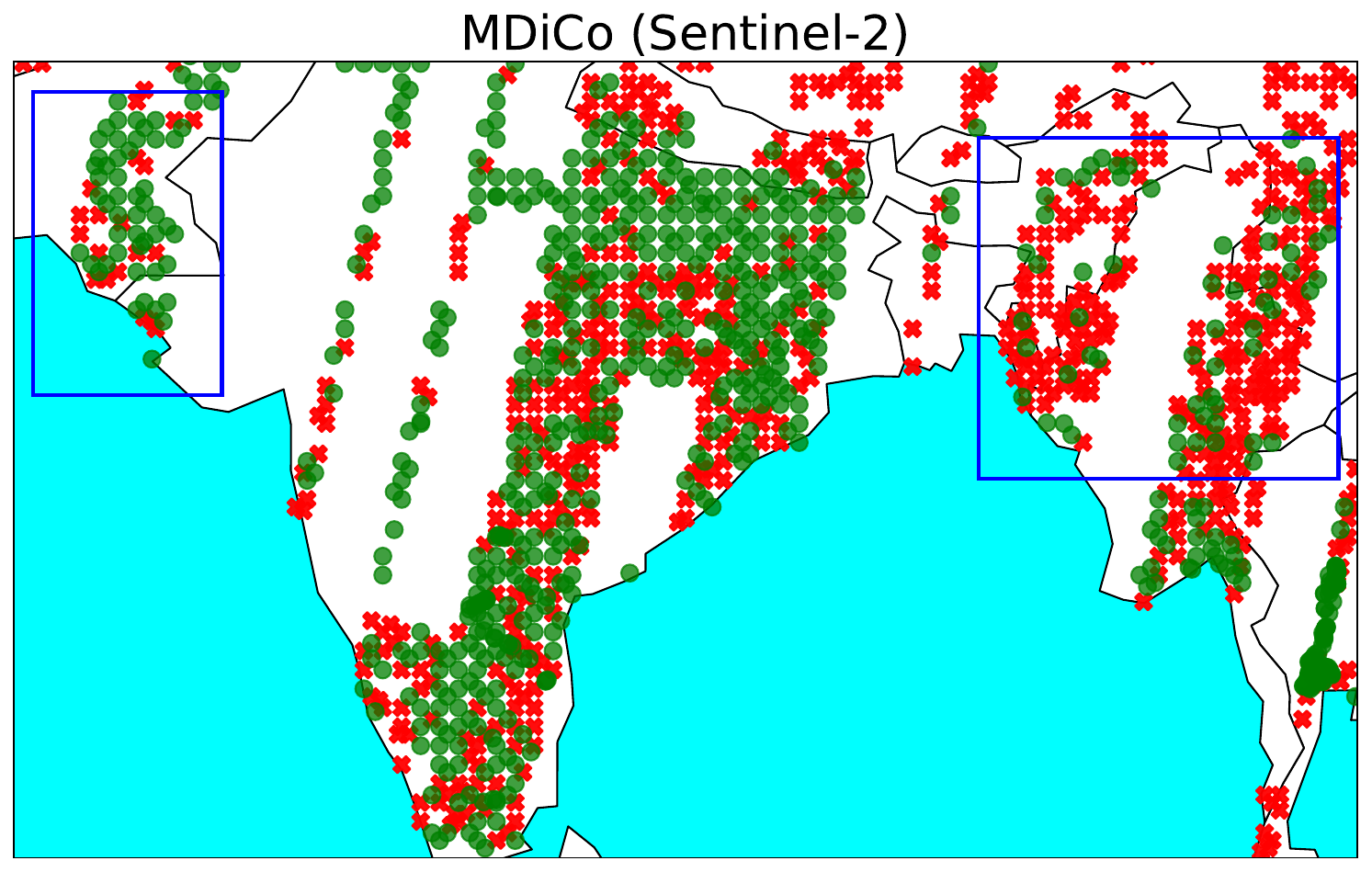}}
    \hfill
	\caption{Prediction map of our \gls{our} method in comparison to the ground truth labels and the Individual baseline using the Sentinel-2 modality. The shown region is in South Asia.}\label{appendix:fig:case3}
\end{figure}

\begin{figure}[!b]
	\centering
    {\includegraphics[width=0.32\textwidth, page=1]{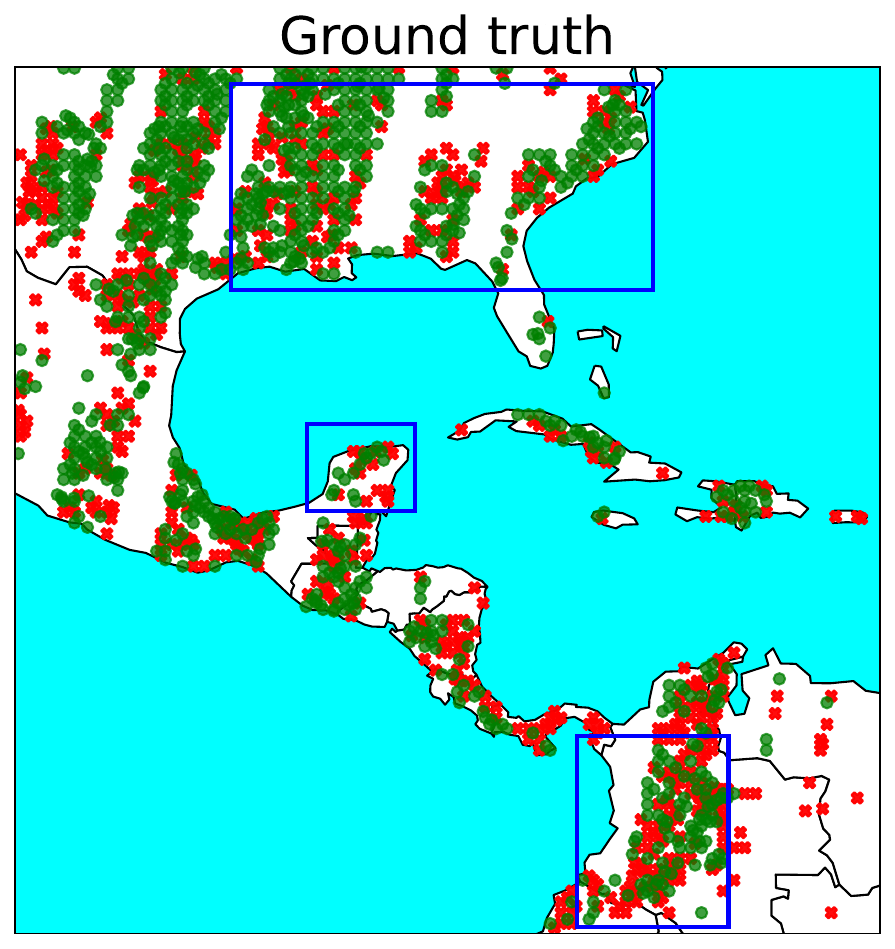}}
    \hfill
    {\includegraphics[width=0.32\textwidth, page=1]{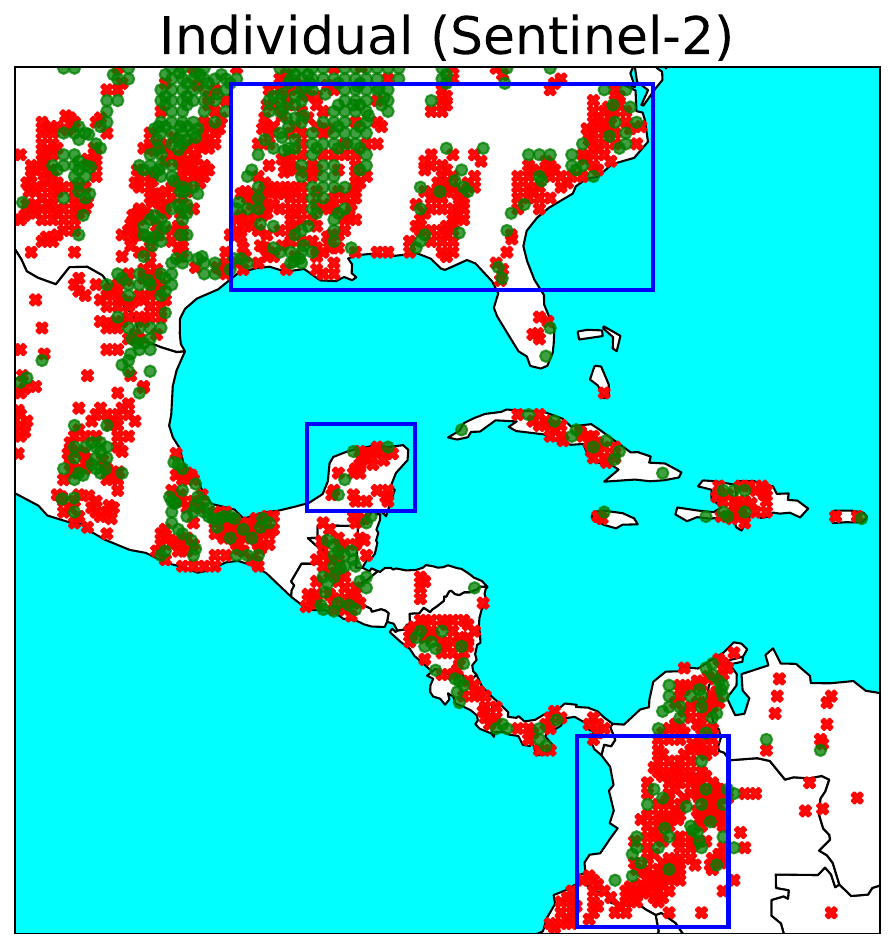}}
    \hfill
    {\includegraphics[width=0.32\textwidth, page=1]{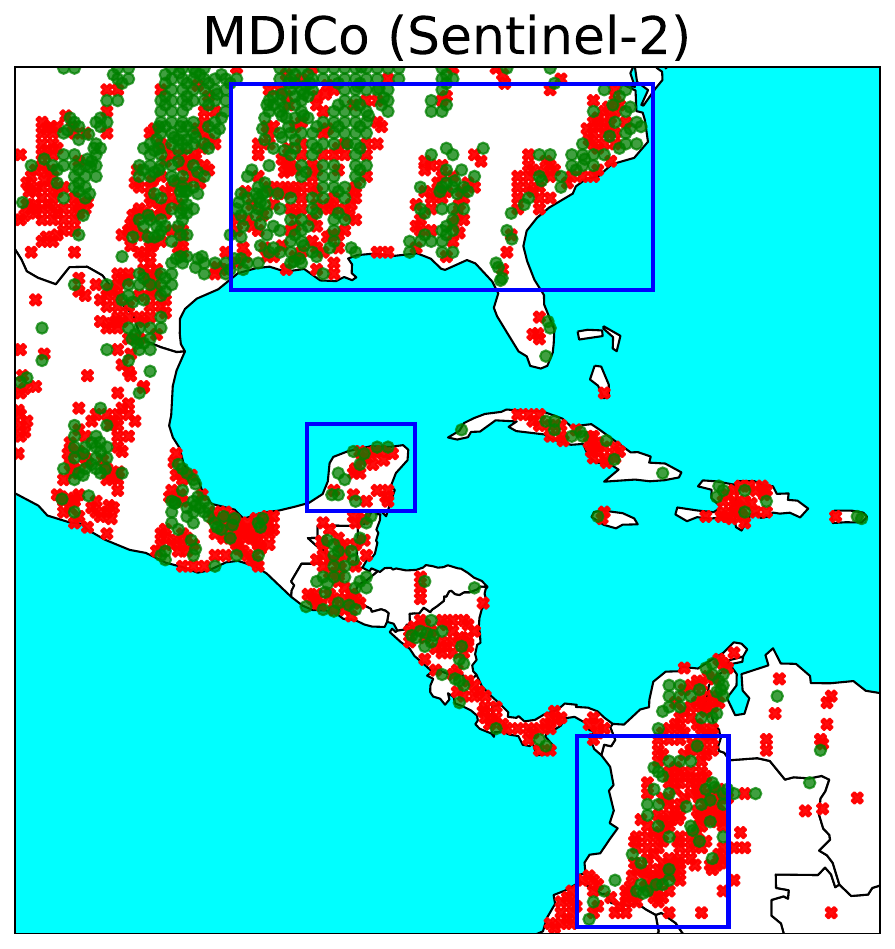}}
    \hfill
	\caption{Prediction map of our \gls{our} method in comparison to the ground truth labels and the Individual baseline using the Sentinel-2 modality. The shown region is in Central America.}\label{appendix:fig:case4}
\end{figure}

We illustrate the prediction map of our approach and the individual baseline using the \gls{cropB} dataset in Fig.~\ref{appendix:fig:case1}-\ref{appendix:fig:case4}. We present three cases of the global (and sparse) binary prediction maps to qualitatively assess the model’s spatial patterns and highlight different regions where there is higher difference.
We observe that the maps generated by our \gls{our} method are more similar to the ground truth in comparison to those produced by the Individual baseline. This highlights the better spatial patterns obtained through the collaboration of multi-modal data during training.

\end{document}